\documentclass[letterpaper, 10 pt, conference]{ieeeconf}  

\usepackage{amsmath}
\usepackage{amssymb}
\usepackage{booktabs}
\usepackage{graphicx}
\usepackage{multirow}
\usepackage{multicol}
\usepackage[table]{xcolor}
\usepackage{adjustbox}
\usepackage{tabularx}
\usepackage{array}
\usepackage[table]{xcolor}
\usepackage{booktabs}
\usepackage{placeins}
\usepackage{cite}
\usepackage{hyperref}
\usepackage{capt-of}
\newcolumntype{T}{>{\hsize=1.15\hsize\centering\arraybackslash}X}
\newcolumntype{P}{>{\hsize=1.25\hsize\centering\arraybackslash}X}
\newcolumntype{O}{>{\hsize=0.60\hsize\centering\arraybackslash}X}
\IEEEoverridecommandlockouts                              

\title{\LARGE \bf
Thermo-VLA: A Thermal-Aware Vision-Language-Action Model \\
with Gated Residual Modulation for Robotic Manipulation
}

\author{Jingxuan Chen, Junshi Luo, Chencheng Dong, Xiaojian Nie, Tongge Huang, Qiang Huang}

\begin{document}

\twocolumn[{%
    \renewcommand\twocolumn[1][]{#1}%
    \maketitle

    {\centering
    \includegraphics[width=\linewidth]{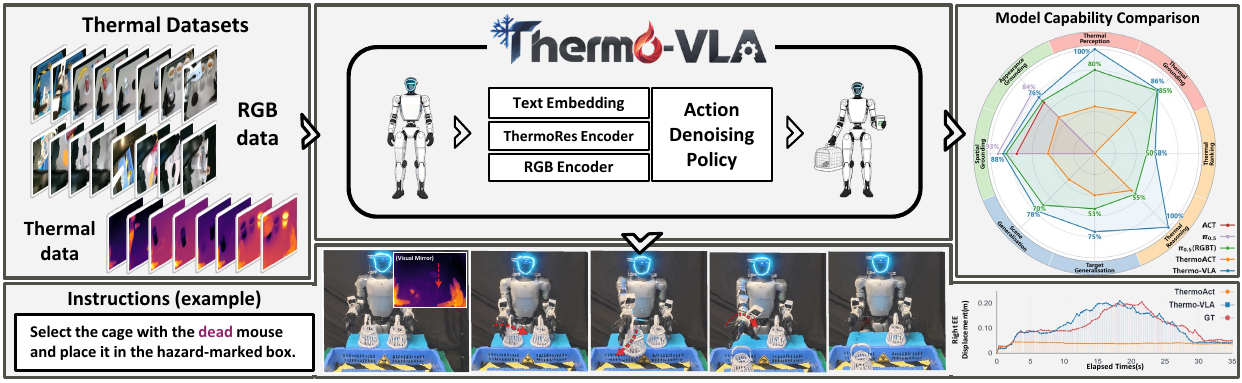}
    \par}

    \captionof{figure}{\textbf{Overview of the Thermo-VLA framework and results.}
    Left: real-world RGB and thermal manipulation data.
    Centre: the Thermo-VLA framework for multimodal encoding and continuous action generation.
    Right: performance across RGB capability, thermal perception, thermal reasoning, and generalisation.
    Bottom: a Task VI example, where Thermo-VLA successfully removes the simulated dead-mouse target and produces arm motion closely aligned with the ground truth.}
    \label{head}

    \vspace{2.5mm}
}]
\thispagestyle{empty}
\pagestyle{empty}


\begin{abstract} 
Recent Vision–Language–Action (VLA) models have begun incorporating thermal sensing, but existing approaches typically treat thermal images as ordinary visual inputs, providing limited support for temperature-aware relational reasoning in manipulation. To address this limitation, we propose Thermo-VLA with a ThermoRes Encoder, which decomposes thermal images into cold and hot attribute representations and injects them into the RGB token through language-conditioned residual modulation, enabling temperature-driven target selection and continuous manipulation. We further construct six real-world robotic tasks on a dual-arm Unitree G1 platform, spanning basic grounding to relational reasoning. Experiments show that Thermo-VLA achieves an average success rate of 0.86 across four thermal-dependent tasks, outperforming a $\pi_{0.5}$\cite{pi0.5} baseline that directly incorporates RGB–thermal observations by 18 percentage points. Under unseen-object and scene variations, Thermo-VLA maintains an average success rate of 0.78, achieving the best generalisation performance among the compared methods. Beyond thermal sensing, we transfer the same residual modulation mechanism to colour attributes and observe consistent improvements in both simulation and real-robot experiments, indicating its potential to generalise beyond thermal attributes. \href{https://j-x-chen.github.io/Thermo-VLA-web/}{Project Website}
\end{abstract}

\section{Introduction}

Instruction-driven robotic manipulation aims to generate continuous actions that are consistent with task goals based on language instructions and environmental observations. In recent years, Vision-Language-Action (VLA) models\cite{pi0,pi0.5,pi0.7,gr00t,rt2,openvla} have substantially advanced language-driven robotic manipulation through heterogeneous pretraining and downstream fine-tuning. However, existing VLAs still rely predominantly on RGB observations for perception. When language instructions involve invisible physical attributes such as temperature,  RGB-based policies are not natively equipped to perceive and model such attributes, limiting both target selection and subsequent action generation.

Recent VLA studies\cite{musevla,thermoact,safenightvla} have begun incorporating physical modalities such as thermal sensing to support manipulation beyond RGB-only perception. Nevertheless, existing thermal research\cite{anythermal,cmx,thermovl,irgpt} remains largely focused on perceptual and understanding tasks, including image fusion, detection and segmentation, grounding, and visual question answering. Thermal manipulation in robotics also remains limited in several aspects. First, existing studies have primarily demonstrated thermal-aware manipulation in relatively straightforward settings like recognising and grasping thermally salient targets, whereas more complex thermal-dependent capabilities, such as relative temperature comparison, conditional selection, and exclusion-based reasoning, remain less explored~\cite{thermoact}. Second, even when temperature-related targets can be accurately identified, improper integration of thermal information can introduce instability, leading to manipulation failures. Finally, existing systems\cite{musevla,thermoact} are predominantly evaluated on fixed-viewpoint, single-arm platforms,  leaving their effectiveness in bimanual egocentric manipulation insufficiently studied.

To address these limitations, we propose Thermo-VLA, an end-to-end thermal-aware VLA framework that integrates RGB, thermal, and language information for temperature-driven manipulation. At its core, Thermo-VLA introduces a dedicated ThermoRes Encoder to incorporate instruction-relevant thermal information while preserving the original RGB representation. Specifically, the ThermoRes Encoder consists of Dual-Branch Attribute Decomposition and Gated Residual Injection. The former decomposes thermal observations into complementary temperature-attribute features, while the latter adaptively injects thermal residuals into RGB tokens under language guidance. As illustrated in the centre of Fig.\ref{head}, the resulting multimodal tokens then condition the action generation module, enabling temperature-driven target selection, grasping, and continuous manipulation.

We systematically evaluate Thermo-VLA on a Unitree G1 bimanual humanoid robot across six real-world manipulation tasks of varying difficulty, covering basic RGB spatial and appearance grounding, thermal perception, thermal grounding, relative temperature ranking, and thermal reasoning. Thermo-VLA outperforms ThermoAct\cite{thermoact} and the $\pi_{0.5}$ baseline that directly incorporates RGB-thermal observations, achieving the best overall performance among the compared methods on the proposed benchmark. As illustrated in the example in Fig.\ref{head}, Thermo-VLA activates the correct right arm and closely follows the ground-truth trajectory, whereas ThermoAct uses the wrong arm and shows little movement in the required right arm. Thermo-VLA also maintains more stable performance under unseen objects, attribute recombination, illumination changes, and scene variations. Ablation studies verify the effectiveness of the ThermoRes Encoder in injecting task-relevant temperature information into visual representations through text-guided residual modulation. Further representation analysis suggests that these residuals provide compact complementary corrections to the RGB representation.

Our main contributions are summarised as follows:
\begin{itemize}
    \item We introduce a real-world benchmark for thermal-aware robotic manipulation, comprising four temperature-related tasks and two RGB-based tasks for evaluating representative VLA methods under RGB and thermal settings.

    \item We propose Thermo-VLA with a ThermoRes Encoder that combines temperature-attribute decomposition and text-conditioned residual injection for temperature-driven manipulation.

    \item Thermo-VLA achieves the best overall performance on the proposed benchmark and demonstrates strong generalisation to unseen objects and scene variations.
\end{itemize}

\section{Related Work}

\subsection{Generalist Robotic Manipulation Methods}
Early end-to-end imitation learning methods\cite{act, diffusionpolicy, vqbet} demonstrated the ability to directly generate continuous actions from visual observations for simple manipulation tasks like grasping and placing. Vision--Language--Action (VLA) models subsequently introduced large-scale vision-language priors into robotic control. RT-2 \cite{rt2} represents actions as text tokens, allowing vision-language knowledge to directly contribute to action prediction, while OpenVLA \cite{openvla} enables efficient adaptation of pretrained vision-language representations to downstream robotic manipulation tasks. The $\pi$ series \cite{pi0,pi0.5,pi0.7} and GR00T series \cite{gr00t} further improve cross-task, cross-embodiment, and long-horizon manipulation through large-scale heterogeneous robot data and generative action modelling. More recently, world-model-based approaches\cite{robodreamer,gr1,gr2,lingbotva,dreamzero} enhance temporal decision-making and closed-loop control by predicting future environment states. Nevertheless, they remain predominantly RGB-based.

\subsection{Multisensory Vision-Language-Action Models}
Recent studies have begun to extend VLA models beyond RGB-centric perception toward richer sensing of the physical world. DepthVLA \cite{depthvla} introduces explicit depth representations to enhance spatial reasoning, while ForceVLA \cite{forcevla} and TA-VLA \cite{tavla} augment VLA models with force/torque feedback. Tactile-VLA \cite{tactilevla}, VTLA \cite{vtla}, and TacVLA \cite{tacvla} leverage tactile sensing for adaptive physical interaction, while Audio-VLA \cite{audiovla} and E-VLA \cite{evla} further extend VLA perception with acoustic and event-based signals, respectively. Beyond individual physical modalities, recent approaches have explored unified multisensory VLA frameworks. Safe-Night VLA\cite{safenightvla} combines thermal and depth information to support temperature-conditioned manipulation and safety-aware control. OmniVLA\cite{omnivla} and MuseVLA\cite{musevla} integrate multiple physical modalities, including thermal, acoustic, and mmWave sensing; the former emphasises unified sensor representations, while the latter dynamically selects required modalities according to the instruction.

\subsection{Thermal Perception, Understanding, and Action}
Recent studies have investigated effective thermal representations and cross-modal fusion\cite{cmx,raccm,freqfusion}. AnyThermal\cite{anythermal} learns a general-purpose thermal backbone through distillation from visual foundation models and transfers it to downstream vision tasks such as segmentation and depth estimation. IRGPT\cite{irgpt} and Thermo-VL\cite{thermovl} further align thermal representations with the language space, supporting infrared image recognition, captioning, and grounding, as well as text-guided RGB-thermal visual question answering and cross-spectral reasoning. More recent work has further incorporated thermal information into robotic manipulation. ThermoAct \cite{thermoact} adopts a hierarchical architecture with a Gemini 2.0 Flash\cite{gemini} planner and a $\pi_0$ executor. However, within the executor, thermal images are converted to pseudocolour and processed by the generic visual encoder, leaving their temperature-specific information insufficiently integrated into action-relevant representations.

\begin{figure*}[t]
\centering
\includegraphics[width=\linewidth]{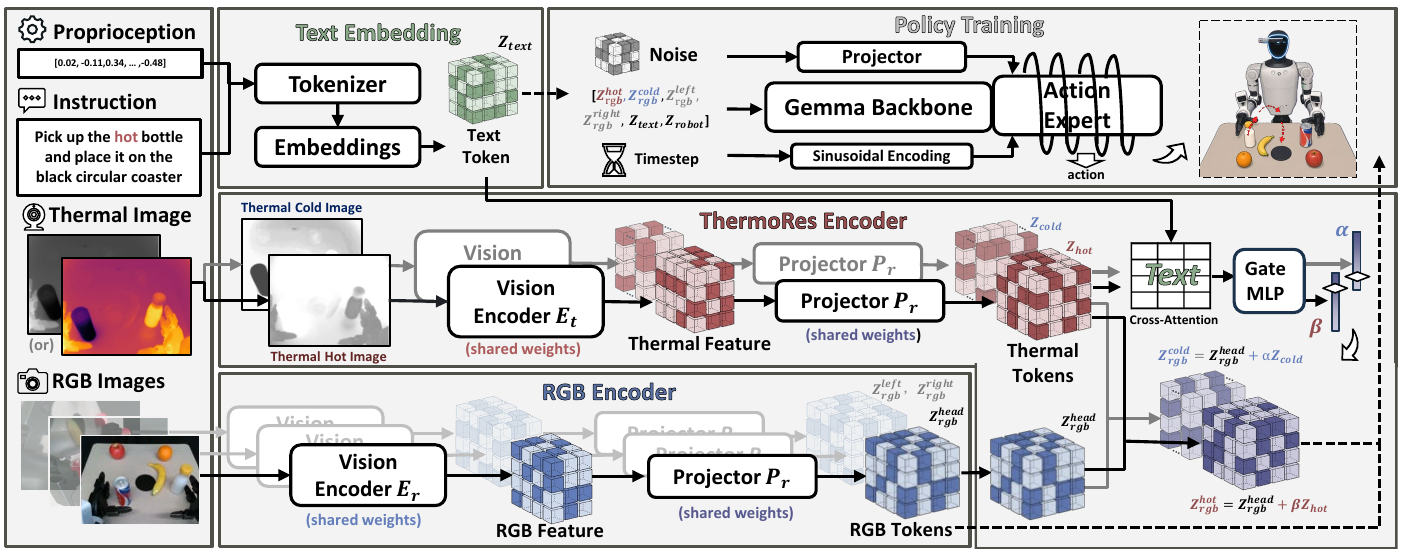}
\caption{\textbf{Overview of the Thermo-VLA framework.} RGB observations are encoded by the RGB visual encoder, while language instructions and robot states are mapped to textual embeddings. ThermoRes Encoder decomposes thermal observations into cold and hot attributes and injects them into RGB tokens through context-gated residuals. The resulting multimodal prefix is used for flow-matching policy training and continuous action generation.}
\vspace{-0.3cm}
\label{pipeline}
\end{figure*}

\section{Thermo-VLA Framework} 
\subsection{Problem Formulation} 
Existing VLA models typically predict continuous robot actions conditioned on visual observations, language instructions, and robot states.  In this work, we introduce Thermo-VLA and use $\pi_{0.5}$ as the baseline model. As illustrated in Fig.~\ref{pipeline}, given RGB observations $x_r$, a thermal observation $x_t$, a language instruction $l$, and the current robot state $s_t$, Thermo-VLA predicts a continuous action chunk of horizon $H$, $A_t=[a_t,\ldots,a_{t+H-1}] \in\mathbb{R}^{H\times 2d_a}$ with conditional distribution
\begin{equation}
A_t\sim\pi_\theta(\cdot\mid x_r,x_t,l,s_t).
\end{equation}
In our setup, $d_a=8$, comprising seven arm-joint dimensions and one hand open/close dimension.

\subsection{ThermoRes Encoder} 
As shown in Fig.\ref{pipeline}, for each RGB observation $x_r^i$, the visual encoder $E_r$ extracts patch-level features, which are projected into the token space by $P_r$: $Z_\text{rgb}^i=P_r(E_r(x_r^i))\in\mathbb{R}^{B\times N\times D}, i\in\{\text{head}, \text{left}, \text{right}\}$, where B, N, and D denote the batch size, number of visual tokens, and token dimensionality, and the left and right RGB observations are captured by wrist-mounted cameras. Since conventional visual encoding is insufficient to explicitly capture temperature attributes in thermal observations, as demonstrated in Sec.~\ref{experiment-1-2}, we further introduce the ThermoRes Encoder, consisting of Dual-Branch Attribute Decomposition and Gated Residual Injection for attribute extraction and context-conditioned integration, respectively.

\subsubsection{Dual-Branch Attribute Decomposition}
Unlike RGB images, whose visual statistics are primarily characterised by colour, texture, and appearance, thermal images mainly reflect scene-dependent temperature distributions. However, cold objects, hot objects, and heat emitted by the robot itself may coexist within a single frame, making it difficult to explicitly identify the required temperature attribute and its response strength from mixed thermal responses. 

To address this issue, we introduce Thermal Dual-Branch Attribute Decomposition. Since thermal responses depend on both the target and the background temperature, the same object may exhibit different absolute intensities across scenes. We first sample multiple patches near the upper image boundary, compute their mean intensities, and take the mode of these values as the background reference $b$ for each frame and calibrate the original thermal image $x_t$ such that the background is mapped to a fixed grayscale intensity $\mu$. The calibrated observation is then decomposed into complementary cold and hot attribute views according to its deviation from the background:
\begin{equation}
\begin{aligned}
k=\operatorname{clip}(&x_t-b+\mu,0,I_{\max}), \\
x_\mathrm{cold}=I_{\max}\frac{\min(k,\mu)}{\mu},& \
x_\mathrm{hot}=I_{\max}\frac{I_{\max}-\max(k,\mu)}{I_{\max}-\mu},
\end{aligned}
\end{equation}
where $I_{\max}$ denotes the upper bound of the grayscale range. This transformation uses the background as a common reference, such that $x_\mathrm{cold}$ and $x_\mathrm{hot}$ emphasise responses below and above the background, respectively. Both branches adopt a shared SigLIP-based thermal encoder $E_t$, with the background close to white and larger temperature deviations appearing darker. The two branches further share the RGB projector $P_r$: $Z_\mathrm{cold}=P_r(E_t(x_\mathrm{cold})), Z_\mathrm{hot}=P_r(E_t(x_\mathrm{hot}))$. This yields temperature attribute tokens with the same dimensionality as the RGB tokens, facilitating subsequent residual injection. 

\subsubsection{Gated Residual Injection}

After Dual-Branch Attribute Decomposition, the cold and hot branches produce separate temperature attribute tokens. Rather than directly concatenating them with other modalities, which increases the burden on downstream multimodal modelling, we introduce Gated Residual Injection to incorporate task-relevant thermal information into the RGB token selectively. It consists of two steps: (1) Text-Gate Cross-Attention estimates the relevance of the cold and hot attributes to the current textual context; and (2) the weighted temperature tokens are injected into the RGB tokens through residual connections.

$ $

\noindent{\normalfont{\textbf{Text-Gate Cross-Attention.}}}
Different task contexts may require different levels of attention to the cold and hot attributes. We therefore use the text token \(Z_{\mathrm{text}}\) as the query and the cold and hot thermal tokens as branch-specific keys and values, producing the cross-attention outputs \(O_c\) and \(O_h\), respectively. The two branches share the attention parameters. We then construct the gating feature from the pooled text token, cross-attention outputs, and their representation differences. Let $u_t:=[\bar{Z}_{\mathrm{text}};\bar{O}_c;\bar{O}_h;\bar{O}_c-\bar{Z}_{\mathrm{text}};\bar{O}_h-\bar{Z}_{\mathrm{text}}]$, where the overbar denotes token-wise pooling. A lightweight Gate MLP directly predicts dynamic weights for the cold and hot branches:
\begin{equation}
[\alpha,\beta]=\operatorname{softmax}\left(\operatorname{MLP}\left(\operatorname{LN}(u_t)\right)\right).
\end{equation}
The contributions of the temperature attributes are thereby adaptively adjusted according to the current textual context.

$ $

\noindent{\normalfont{\textbf{Residual Injection.}}}
The RGB token primarily provides the appearance, spatial grounding, and visual semantics required for action prediction, while thermal observations provide temperature cues unavailable from RGB. Inspired by the residual learning principle of ResNet\cite{resnet}, we preserve the RGB token as an identity path and inject the temperature attributes using the predicted dynamic weights:
\begin{equation}
Z_\mathrm{rgb}^\mathrm{cold}=Z_\mathrm{rgb}^\mathrm{head}+\alpha Z_\mathrm{cold},\
Z_\mathrm{rgb}^\mathrm{hot}=Z_\mathrm{rgb}^\mathrm{head}+\beta Z_\mathrm{hot}.
\end{equation}
As the RGB and thermal branches are mapped into a compatible feature space, $Z_\mathrm{rgb}^\mathrm{head}$, $Z_\mathrm{cold}$, and $Z_\mathrm{hot}$ share the same token layout and token dimensionality, enabling token-wise residual modulation of the RGB representation. Independently modulated by the textual context, $Z_\mathrm{rgb}^\mathrm{cold}$ and $Z_\mathrm{rgb}^\mathrm{hot}$ are combined with the remaining conditioning tokens to form the multimodal prefix for subsequent action generation.

\begin{table}[t]
\centering
\caption{\textbf{Dataset specifications.} The six real-world tasks, their key challenges, and the number of expert demonstrations.}
\label{tab:dataset_specs}

\setlength{\tabcolsep}{3pt}
\renewcommand{\arraystretch}{1.1}

\begin{adjustbox}{max width=\columnwidth}
\begin{tabular}{@{}lllc@{}}
\toprule
\textbf{No.} & \textbf{Task Name} & \textbf{Difficulty} & \textbf{Episodes} \\
\midrule
\textbf{I}   & Position-guided bottle selection       & Spatial Grounding    & 156 \\
\textbf{II}  & Temperature-guided bottle selection    & Thermal Perception   & 303 \\
\textbf{III} & Identity-guided object selection        & Appearance Grounding & 236 \\
\textbf{IV}  & Temperature-guided object selection     & Thermal Grounding    & 340 \\
\textbf{V}   & Temperature-aware coffee service        & Thermal Ranking      & 173 \\
\textbf{VI}  & Abnormal-animal cage removal          & Thermal Exclusion  & 99 \\
\bottomrule
\end{tabular}
\end{adjustbox}
\vspace{-0.3cm}
\end{table}
\subsection{Action Generation}

After Gated Residual Injection, the enhanced visual tokens are combined with the text token $Z_{text}$, which is formed from the language instruction $l$ and the discretised robot state $s_t$ by tokenisation and embedding. Together, they form the temperature-conditioned prefix $Z_T:=[ Z_\mathrm{rgb}^\mathrm{cold}; Z_{rgb}^\mathrm{hot}; Z_\mathrm{rgb}^\mathrm{left}; Z_\mathrm{rgb}^\mathrm{right}; Z_\mathrm{text}]$. The resulting prefix $Z_T$ is then processed by the Gemma-based multimodal backbone and provided as conditioning information to the action expert. We retain the base model's flow-matching action generation mechanism. Given the action chunk $A_t$ and Gaussian noise $\epsilon\sim\mathcal{N}(0,I)$, we construct $x_\tau=(1-\tau)A_t+\tau\epsilon,$ where the flow-matching time step $\tau$ is encoded using sinusoidal encoding as the temporal condition. During inference, the action variable is initialised from Gaussian noise and evolved along the predicted velocity field to generate a continuous action chunk.
\section{Experiments}
\begin{table}[t]
\centering
\caption{\textbf{Success rates on the six real-world manipulation tasks.} Best results are highlighted in bold.}
\label{result_table1}

\setlength{\tabcolsep}{1.5pt}
\renewcommand{\arraystretch}{1.15}

\begin{adjustbox}{max width=\columnwidth}
\begin{tabular}{@{}l@{\hspace{1pt}}l@{\hspace{-2.5pt}}c>{\hspace{4pt}}cc>{\hspace{-2.5pt}}cc@{}}
\toprule

\multirow{2}{*}{\textbf{Task}}
& \multirow{2}{*}{\ \textbf{Instruction}}
& \multicolumn{2}{c}{\textbf{RGB}}
& \multicolumn{3}{c}{\textbf{RGBT}} \\
\cmidrule(lr){3-4}\cmidrule(lr){5-7}

&
& \scriptsize\textbf{ACT}
& \scriptsize\textbf{$\pi_{0.5}$}
& \scriptsize\textbf{$\pi_{0.5}$\tiny(RGBT)}
& \fontsize{6}{7}\selectfont\textbf{ThermoAct}
& \scriptsize\textbf{Ours} \\
\midrule

\multicolumn{7}{l}{\textit{\textbf{Task I: Position-guided bottle selection}}} \\
& Left bottle $\rightarrow$ black coaster
& 0.75 & 0.95 & 0.95 & 0.80 & \textbf{0.95} \\
& Right bottle $\rightarrow$ black coaster
& 0.75 & 0.90 & 0.75 & 0.10 & 0.80 \\
\rowcolor{gray!15}
\textbf{Avg.} & 
& 0.75 & \textbf{0.93} & 0.85 & 0.45 & 0.88 \\
\midrule

\multicolumn{7}{l}{\textit{\textbf{Task II: Temperature-guided bottle selection}}} \\
& Cold bottle $\rightarrow$ black coaster
& / & / & 0.80 & 0.45 & \textbf{1.00} \\
\midrule

\multicolumn{7}{l}{\textit{\textbf{Task III: Identity-guided object selection}}} \\
& Milk $\rightarrow$ yellow coaster
& 0.65 & 0.85 & 0.75 & 0.55 & 0.80 \\
& Milk $\rightarrow$ white coaster
& 0.75 & 0.85 & 0.70 & 0.45 & \textbf{0.85} \\
& Coke $\rightarrow$ yellow coaster
& 0.65 & 0.80 & 0.70 & 0.45 & 0.70 \\
& Coke $\rightarrow$ white coaster
& 0.70 & 0.85 & 0.70 & 0.50 & 0.70 \\
\rowcolor{gray!15}
\textbf{Avg.} &
& 0.69 & \textbf{0.84} & 0.71 & 0.49 & 0.76 \\
\midrule

\multicolumn{7}{l}{\textit{\textbf{Task IV: Temperature-guided object selection}}} \\
& Hot object $\rightarrow$ yellow coaster
& / & / & 0.85 & 0.60 & 0.90 \\
& Hot object $\rightarrow$ white coaster
& / & / & 0.90 & 0.55 & 1.00 \\
& Cold object $\rightarrow$ yellow coaster
& / & / & 0.75 & 0.50 & 0.70 \\
& Cold object $\rightarrow$ white coaster
& / & / & 0.90 & 0.55 & 0.85 \\
\rowcolor{gray!15}
\textbf{Avg.} &
& \textbf{/} & \textbf{/} & 0.85 & 0.55 & \textbf{0.86} \\
\midrule

\multicolumn{7}{l}{\textit{\textbf{Task V: Temperature-aware coffee service}}} \\
& Sugar $\rightarrow$ Less ice
& / & / & 0.50 & 0.00 & 0.60 \\
& Sugar $\rightarrow$ Extra ice
& / & / & 0.50 & 0.00 & 0.55 \\
\rowcolor{gray!15}
\textbf{Avg.} &
& / & / & 0.50 & 0.00 & \textbf{0.58} \\
\midrule

\multicolumn{7}{l}{\textit{\textbf{Task VI: Abnormal-animal cage removal}}} \\
& Dead-mouse cage $\rightarrow$ hazard box
& / & / & 0.55 & 0.50 & \textbf{1.00} \\
\bottomrule

\end{tabular}
\end{adjustbox}
\vspace{-0.2cm}
\end{table}

\subsection{Comparative Experiments}
\begin{figure*}[t]
\centering
\includegraphics[width=0.98\linewidth]{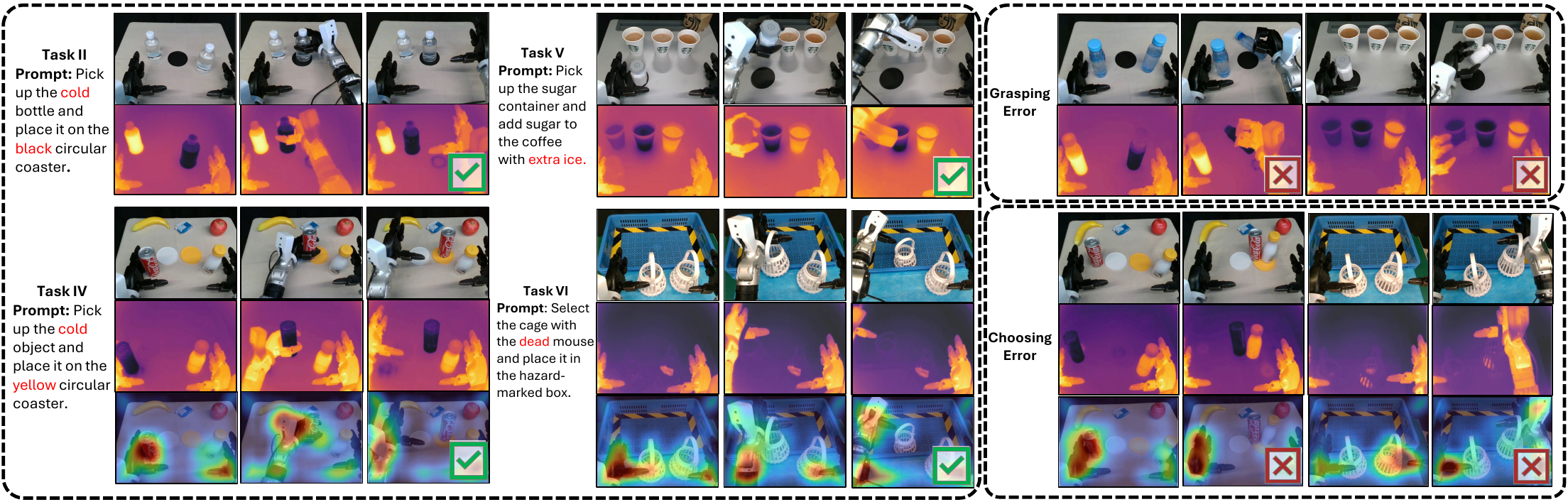}
\caption{\textbf{Representative successful and failed executions.} Left: successful executions of Thermo-VLA. Right: representative failure cases of competing RGB--thermal methods. The bottom row shows the corresponding attention maps.}
\label{result1}
\vspace{-0.1cm}
\end{figure*}
\subsubsection{Task Description}

We design six real-world robotic manipulation tasks, with rigidly mounted head RGB and thermal cameras at a fixed relative pose; detailed hardware specifications are provided in Appendix II on the project website. Tasks I and III are RGB-only baseline tasks, in which the robot selects and manipulates targets based on left-right position and object identity, respectively. Tasks II and IV replace the selection criterion with temperature attributes, extending the evaluation to temperature-dependent manipulation. Tasks V and VI further adopt settings inspired by practical applications in drink service and laboratory handling. Specifically, Task V requires the robot to distinguish among iced, mildly cool, and hot states and add sugar to the specified coffee, while Task VI requires the robot to identify and transfer a target exhibiting an abnormal thermal response(e.g., the dead mouse). Table~I summarises the task settings, key challenges, and the 1,307 expert demonstrations collected for training. During evaluation, Tasks I--IV are considered successful only when the correct target is placed in the correct coaster; Task V requires pouring sugar over the specified coffee, and Task VI requires transferring the cage containing the mouse without thermal response to the hazard-marked box. Each instruction is evaluated over 20 trials. After each trial, an operator wearing thermal-insulating gloves resets the objects, with approximately matched positions across methods for the same trial, while object layouts vary across trials. Liquid weights are kept consistent within each test group.

\subsubsection{Comparison Results}
\label{experiment-1-2}
We compare the RGB-only methods ACT~\cite{act} and $\pi_{0.5}$~\cite{pi0.5} with three RGB--thermal methods: $\pi_{0.5}$ (RGBT), ThermoAct~\cite{thermoact}, and Thermo-VLA. All baselines follow the hyperparameter settings in their original papers. For $\pi_{0.5}$ (RGBT), the thermal image is directly introduced as an additional visual observation. On the RGB-only Tasks I and III, both $\pi_{0.5}$ and Thermo-VLA maintain strong visual grounding and manipulation performance, although Thermo-VLA is slightly lower than the original $\pi_{0.5}$. Nevertheless, Thermo-VLA shows the smallest degradation among the RGB--thermal methods, suggesting that the additional thermal modality causes greater interference with RGB-based manipulation in the competing methods. On the temperature-dependent Tasks II and IV, Thermo-VLA achieves success rates of 1.00 and 0.86, respectively, showing only a modest advantage over $\pi_{0.5}$ (RGBT) on these relatively simple thermal tasks, while ThermoAct exhibits a more evident unilateral action bias. As the tasks require thermal reasoning, the advantage of Thermo-VLA becomes more apparent. On Task V, which involves relative temperature reasoning, Thermo-VLA achieves a success rate of 0.58, outperforming $\pi_{0.5}$ (RGBT). The difference becomes most pronounced on Task VI, where Thermo-VLA succeeds in all trials, while the other RGB-thermal methods achieve success rates of only around 0.5 and frequently select the wrong target. Fig.~\ref{result1} also illustrates representative failure cases. ThermoAct occasionally suffers from unstable grasping, while $\pi_{0.5}$ (RGBT) may either manipulate both objects simultaneously or incorrectly select the salient heated object. In contrast, the attention maps show that Thermo-VLA focuses on the non-heated target rather than the dominant heat sources in Task VI.

Overall, Thermo-VLA remains competitive on RGB-only manipulation despite performing slightly below $\pi_{0.5}$, which may reflect the large-scale RGB pretraining of $\pi_{0.5}$ and the absence of comparable thermal pretraining for Thermo-VLA. More importantly, Thermo-VLA achieves clear advantages on temperature-dependent tasks, more consistently exploiting thermal evidence for target selection and higher-level temperature reasoning.

\subsection{Generalization Experiments}
\subsubsection{Task Description}

The generalisation experiments examine whether the model can transfer its learned temperature-grounding capability to unseen objects and scenes. We consider four settings: (1) reassigning temperature states to seen objects, such as changing hot milk to cold milk and cold cola to warm cola, to test whether object appearance and temperature are rigidly coupled; (2) replacing training targets with unseen containers or object categories, including a cold water bottle, a wider milk bottle, an energy drink bottle, and coffee cups from unseen brands; (3) introducing red illumination to perturb RGB appearance; and (4) replacing the tablecloth with two unseen variants, including a more realistic coffee-shop-style setting. For Task VI, we only replace the simulated mouse with a smaller unseen target to reduce the visible thermal region. Since laboratory environments are typically more visually stable, no additional illumination or tablecloth variations are introduced for this task. All methods follow the same evaluation protocol as in the main experiments.

\begin{table*}[t]
\centering
\caption{\textbf{Generalisation performance under target and scene variations.}
For ''Seen Objects'', temperature attributes are reassigned to previously seen objects. Results are reported as success rates.}
\label{result_table2}

\footnotesize
\renewcommand{\arraystretch}{1.15}
\setlength{\tabcolsep}{1pt}

\begin{tabularx}{\textwidth}{@{}cTPOTPOTPOTPO@{}}
\toprule

& \multicolumn{6}{c}{\textbf{Target Generalisation}}
& \multicolumn{6}{c}{\textbf{Scene Generalisation}} \\
\cmidrule(lr){2-7}
\cmidrule(lr){8-13}

\textbf{Task}
& \multicolumn{3}{c}{\textbf{Seen Objects}}
& \multicolumn{3}{c}{\textbf{Unseen Objects}}
& \multicolumn{3}{c}{\textbf{Illumination Change}}
& \multicolumn{3}{c}{\textbf{Unseen Tablecloth}} \\
\cmidrule(lr){2-4}
\cmidrule(lr){5-7}
\cmidrule(lr){8-10}
\cmidrule(lr){11-13}

& {\scriptsize\textbf{ThermoAct}}
& {\footnotesize\textbf{$\pi_{0.5}$}\fontsize{6}{7}\selectfont\textbf{(RGBT)}}
& {\footnotesize\textbf{Ours}}

& {\scriptsize\textbf{ThermoAct}}
& {\footnotesize\textbf{$\pi_{0.5}$}\fontsize{6}{7}\selectfont\textbf{(RGBT)}}
& {\footnotesize\textbf{Ours}}

& {\scriptsize\textbf{ThermoAct}}
& {\footnotesize\textbf{$\pi_{0.5}$}\fontsize{6}{7}\selectfont\textbf{(RGBT)}}
& {\footnotesize\textbf{Ours}}

& {\scriptsize\textbf{ThermoAct}}
& {\footnotesize\textbf{$\pi_{0.5}$}\fontsize{6}{7}\selectfont\textbf{(RGBT)}}
& {\footnotesize\textbf{Ours}} \\
\midrule

\textbf{II}
& / & / & /
& 0.55 & 0.85 & 1.00
& 0.50 & 0.70 & 0.90
& 0.45 & 0.80 & 0.90 \\

\textbf{IV}
& 0.50 & 0.10 & 0.60
& 0.45 & 0.70 & 0.73
& 0.55 & 0.85 & 0.83
& 0.58 & 0.85 & 0.88 \\

\textbf{V}
& / & / & /
& 0.00 & 0.50 & 0.58
& 0.00 & 0.50 & 0.55
& 0.00 & 0.50 & 0.58 \\
\textbf{VI}
& / & / & /
& 0.50 & 0.50 & 0.85
& / & / & /
& / & / & / \\

\rowcolor{gray!15}
\textbf{Avg.}
& 0.50 & 0.10 & \textbf{0.60}
& 0.38 & 0.64 & \textbf{0.79}
& 0.35 & 0.68 & \textbf{0.76} 
& 0.34 & 0.72 & \textbf{0.79}\\

\bottomrule
\end{tabularx}

\end{table*}

\begin{figure}[t]
\centering
\includegraphics[width=\linewidth]{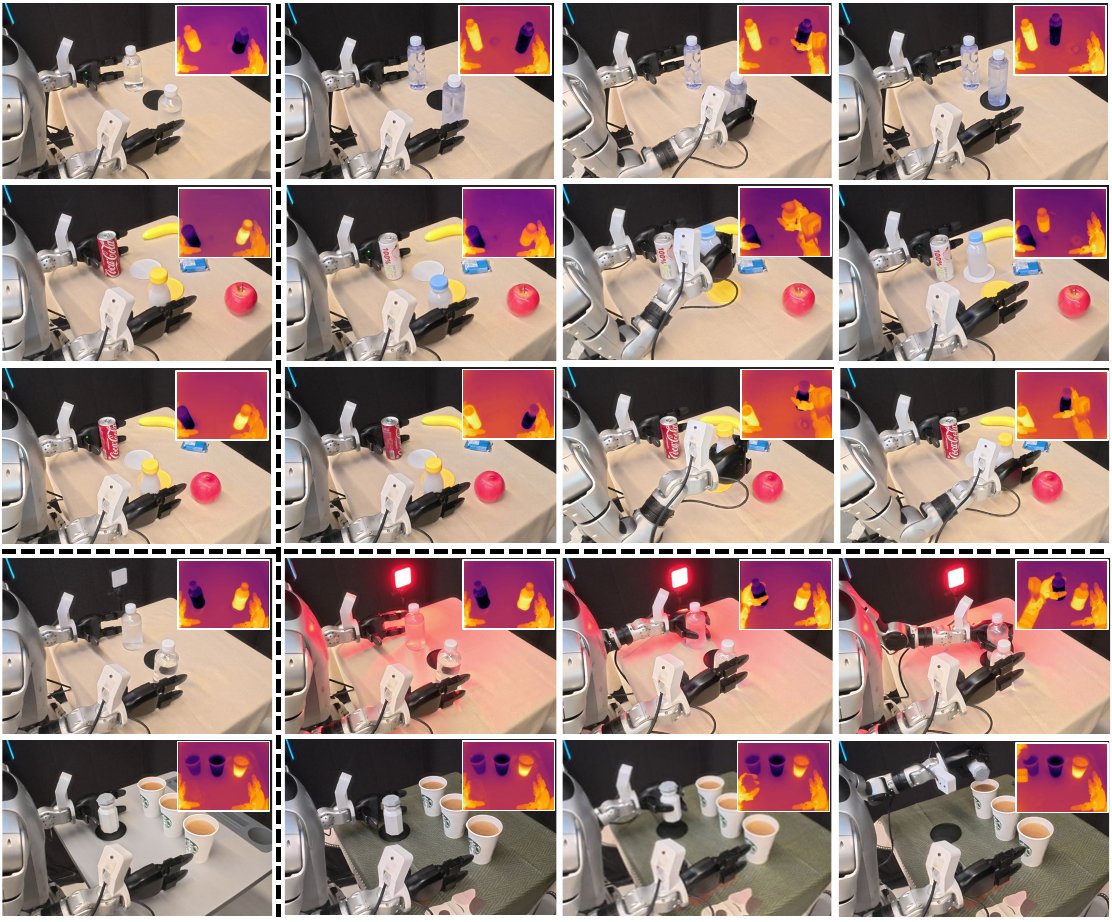}
\caption{\textbf{Generalisation settings and representative successful executions of Thermo-VLA.} The first two columns show the original and modified scenes, with only the corresponding generalisation factor changed; the remaining columns show successful executions. From top to bottom: unseen bottle, unseen objects, temperature reassignment, illumination change, and unseen tablecloth.}
\label{result2}
\vspace{-0.1cm}
\end{figure}

\subsubsection{Generalization Results}

Using the same checkpoints as in the main experiments, we compare ThermoAct, $\pi_{0.5}$ (RGBT), and Thermo-VLA. Thermo-VLA achieves the best overall performance across the four generalisation settings. As shown in Table \ref{result_table2} and Fig.\ref{result2}, its average success rate on unseen objects reaches 0.79, exceeding $\pi_{0.5}$ (RGBT) and ThermoAct by 0.15 and 0.41, respectively, while maintaining an overall advantage under attribute reassignment, unseen tablecloths, and illumination changes. In detail, under temperature-attribute reassignment in Task IV, the success rate of Thermo-VLA drops to 0.60, indicating that swapping the temperature states of seen objects remains challenging. In comparison, $\pi_{0.5}$ (RGBT) is more strongly affected by the appearance--temperature associations learned during training. A different trend is observed in the unseen-object setting. In Task II, Thermo-VLA maintains perfect performance on the wider unseen water bottle, whereas its performance decreases in Task VI as the target and visible thermal region become smaller. This suggests that, beyond object novelty, the spatial extent and morphology of thermal evidence also affect generalisation. Scene variations have a comparatively limited effect on Thermo-VLA. Its average success rates under illumination changes and unseen tablecloths are 0.76 and 0.79, respectively. This indicates limited sensitivity to RGB colour changes and tabletop backgrounds. In addition, Task V changes only slightly across these settings, suggesting that this trend is consistent with relative-temperature discrimination remaining a bottleneck rather than scene variation.

\subsection{Ablation Studies}

We conduct ablation studies on Tasks II, IV, and VI together with their corresponding generalisation settings, covering temperature-based selection, cross-object thermal grounding, scene perturbations, and small-scale thermal targets. Removing Gated Residual Injection leads to the largest performance drop on Task VI$^{\mathrm{Object}}$(smaller unseen target), where the success rate decreases from 0.85 to 0.60. This suggests that text-conditioned residual modulation is particularly important when thermal evidence is weak or spatially limited. Replacing the dual-branch attribute views with direct pseudocoloured thermal input reduces performance under scene perturbations, from 0.90 to 0.85 on Task II\(^{Light}\) and from 0.88 to 0.80 on Task IV\(^{Table}\). This supports the robustness of the proposed temperature-attribute representation. When both components are removed, the model is close to $\pi_{0.5}$ (RGBT), and the advantage introduced by the ThermoRes Encoder is correspondingly reduced.

Finally, we force the thermal residual to zero to examine its role within the policy. Under this setting, Thermo-VLA achieves success rates close to 0.5 across temperature-dependent tasks. The model can still perform basic reaching, grasping, and placing motions, but can no longer reliably select targets according to temperature information. This indicates that the original visual-action representation retains basic manipulation, whereas the thermal residual mainly provides temperature-conditioned target modulation.

\begin{table}[t]
\centering
\caption{Ablation Study}
\label{ablation}

\footnotesize
\setlength{\tabcolsep}{3pt}
\renewcommand{\arraystretch}{1.1}

\begin{tabular}{@{}l|cc|c@{\hspace{1pt}}c@{\hspace{1pt}}c@{}}
\toprule
\scriptsize\textbf{Method\textbackslash Task}
& \scriptsize Task II
& \scriptsize Task IV
& \scriptsize Task II\scalebox{0.90}{$^{Light}$}
& \scriptsize Task IV\scalebox{0.90}{$^{Table}$}
& \scriptsize Task VI\scalebox{0.90}{$^{Object}$} \\
\midrule
\textbf{\scriptsize Thermo-VLA (Ours)}
& 1.00 & 0.86 & 0.90 & 0.88 & 0.85 \\

\textbf{\scriptsize w/o Gated Res. Inj.}
& 1.00 & 0.80 & 0.90 & 0.85 & 0.60 \\

\textbf{\scriptsize w/o Dual-Branch Split}
& 1.00 & 0.85 & 0.85 & 0.80 & 0.85 \\

\textbf{\scriptsize w/o Both}
& 0.80 & 0.85 & 0.70 & 0.85 & 0.50 \\
\midrule
\textbf{\scriptsize Thermal Res. = 0}
& 0.50 & 0.45 & 0.50 & 0.45 & 0.50 \\

\bottomrule
\end{tabular}
\vspace{-0.1cm}
\end{table}

\section{Discussion}

From a more general perspective, the residual formulation in our framework can be expressed as
\begin{equation}
y=x+f(a;l),
\end{equation}
where $x$ denotes the original RGB visual representation, $ a$ denotes an attribute observation associated with the current visual observation, $l$ denotes the language condition, and $f(\cdot)$ denotes the resulting attribute-conditioned residual. Spectral analysis shows that most residual variation is concentrated in a small number of principal components, exhibiting an approximately low-rank structure. According to the Eckart--Young--Mirsky theorem~\cite{eym}, such residuals can be well approximated within a low-dimensional subspace; detailed analysis is provided in Appendix VII on the project website. Intuitively, this indicates that attribute modulation is concentrated along a small set of dominant directions rather than arbitrary high-dimensional perturbations. We further investigate whether these directions are thermal-specific or attribute-dependent by transferring the same residual mechanism to pure-RGB colour-conditioned tasks.

\subsection{RGBRes Encoder}
\begin{figure}[t]
\centering
\includegraphics[width=\linewidth]{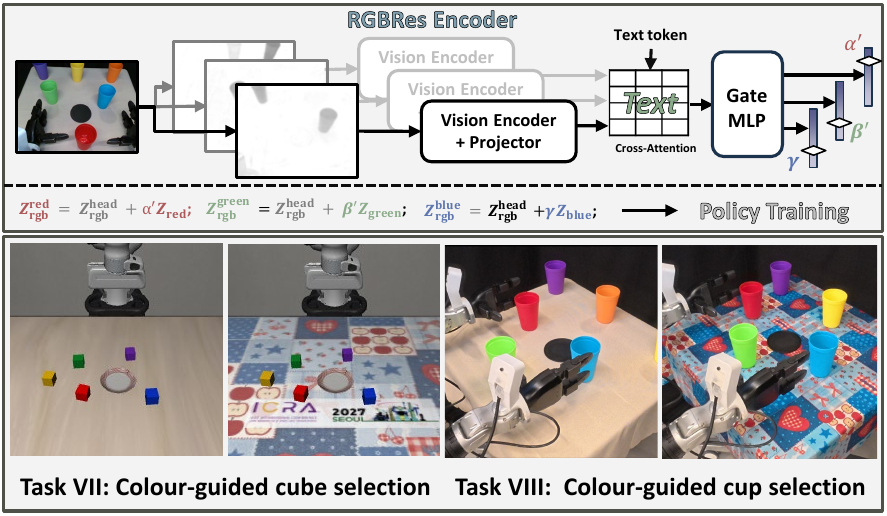}
\caption{\textbf{RGBRes Encoder and representative scenes from the additional colour-conditioned tasks.} Task VII is shown in simulation and Task VIII in the real world.}
\label{rgb}
\vspace{-0.2cm}
\end{figure}

As illustrated in Fig.\ref{rgb}, RGBRes preserves the overall residual architecture of the ThermoRes Encoder while replacing temperature attributes with colour attributes. Given a normalised RGB image, we construct three colour attribute maps from red, green, and blue channels, respectively. Following the same attribute rendering scheme as ThermoRes, these maps are converted into the corresponding black-on-white attribute views $x_j^{col}, j\in\{\mathrm{red},\mathrm{green},\mathrm{blue}\}$. The three colour attributes share the same colour visual encoder $E_{\mathrm{col}}$ and projector $P_{\mathrm{col}}$, yielding $Z_j=P_{\mathrm{col}}\!\left(E_{\mathrm{col}}\!\left(x_j^{col}\right)\right)$. The remaining cross-modal pooling and gating operations follow ThermoRes. The Text-Gate directly predicts three dynamic weights, $\alpha'$, $\beta'$, and $\gamma$, for the red, green, and blue branches, respectively. The corresponding colour attributes are then injected into the original RGB tokens as weighted additive residuals:
\begin{equation}
\begin{aligned}
Z^\mathrm{red}_\mathrm{rgb}&=Z_\mathrm{rgb}^\mathrm{head}+\alpha' Z_\mathrm{red}, \\
Z^\mathrm{green}_\mathrm{rgb}&=Z_\mathrm{rgb}^\mathrm{head}+\beta' Z_\mathrm{green},\ \\
Z^\mathrm{blue}_\mathrm{rgb}&=Z_\mathrm{rgb}^\mathrm{head}+\gamma Z_\mathrm{blue}.
\end{aligned}
\end{equation}
The contributions of the three colour attributes are thereby adaptively adjusted according to the current textual context. 

\subsection{Additional Experiments}
To examine whether this structure extends to other attributes, we apply RGBRes to four pure-RGB colour-conditioned tests in simulation and the real world, as shown in Fig.~\ref{rgb}. Task VII is built on LIBERO with a Franka manipulator, where the robot is instructed to pick a specified red, blue, or green cube from five coloured cubes and place it on a plate. Each of the three target-colour subtasks contains 50 training trajectories and 50 testing trajectories. Task VII$^{\mathrm{Table}}$ replaces the original tabletop with a visually complex tablecloth containing dense red and blue patterns to evaluate robustness against background colour interference. Task VIII is a real-world bimanual manipulation task in which the robot selects either a red or blue cup from six coloured cups and places it on a black coaster. The training set contains 124 demonstrations. Task VIII$^{\mathrm{Table}}$ uses a similar red-blue patterned tablecloth during evaluation. As shown in Table~\ref{rgb_table}, RGBRes consistently improves performance across all four tests, increasing the average success rate from $0.572$ to $0.775$. The improvement remains consistent across simulation, real-world manipulation, and visually complex backgrounds. 

Building on (1) the residual-zero ablation, (2) the approximately low-rank structure revealed by spectral analysis, and (3) the colour-transfer experiments, we arrive at a functional interpretation of the proposed mechanism: (1) task-relevant attribute information is primarily carried by the residual; (2) this information is concentrated along a small set of dominant correction directions; and (3) this low-dimensional directional structure may generalise to represent a broader range of task-relevant physical attributes. Together, these findings suggest that the residual-form representation can be functionally separated into a base component that preserves generic vision--action capability and an additive residual component that carries additional task-relevant physical attributes.

\begin{table}[t]
\centering
\caption{\textbf{Success rates on the additional colour-conditioned manipulation tasks.}}
\label{rgb_table}

\footnotesize
\setlength{\tabcolsep}{2pt}
\renewcommand{\arraystretch}{1.1}

\begin{tabular}{@{}l@{}cccc@{}}
\toprule
\textbf{Method}
& Task VII
& Task VII\scalebox{0.90}{$^{Table}$}
& Task VIII
& Task VIII\scalebox{0.90}{$^{Table}$} \\
\midrule

\textbf{$\pi_{0.5}$}
&  0.707 & 0.307 & 0.600 & 0.675 \\

\textbf{$\pi_{0.5}$} \footnotesize + RGBRes Encoder
&  0.873 & 0.453 & 0.900 & 0.875 \\
\bottomrule
\end{tabular}
\vspace{-0.2cm}
\end{table}

\section{Conclusion}
We propose Thermo-VLA, which introduces a ThermoRes Encoder to inject temperature attributes from thermal images into RGB visual representations as text-conditioned residuals, enabling temperature-driven target selection and continuous manipulation while preserving fundamental vision–action capabilities. We establish a real-world benchmark on the Unitree G1 platform, covering RGB-only grounding, temperature-dependent manipulation, and thermal reasoning. Experiments show that Thermo-VLA achieves superior performance across the benchmark and diverse generalisation settings. Beyond thermal sensing, our results support low-dimensional attribute-conditioned residual modulation as a broader representation principle for augmenting generic vision--action representations with additional physical attributes.

\noindent{\normalfont{\textbf{Limitations and Future Work.}}}
A natural next step is to further explore the residual formulation with other modalities, for example, creating a DepthRes Encoder with complementary near/far attribute maps, to further support residual modulation as a general mechanism for attribute injection. At the same time, we will return to Thermo-VLA itself and address its remaining limitations. We plan to pretrain the thermal encoder on larger-scale thermal data to better preserve RGB-only capabilities, and augment the current cold/hot decomposition with temperature-magnitude information, enabling the model to distinguish finer differences such as cold versus cool objects in Task V.

\bibliographystyle{IEEEtran}
\bibliography{references}

@inproceedings{act,
  title     = {Learning Fine-Grained Bimanual Manipulation with Low-Cost Hardware},
  author    = {Zhao, Tony Z. and Kumar, Vikash and Levine, Sergey and Finn, Chelsea},
  booktitle= {Proceedings of Robotics: Science and Systems (RSS)},
  year      = {2023},
  month     = {July},
  doi       = {10.15607/RSS.2023.XIX.016},
  publisher = {Robotics: Science and Systems Foundation}
}

@inproceedings{diffusionpolicy,
  title     = {Diffusion Policy: Visuomotor Policy Learning via Action Diffusion},
  author    = {Chi, Cheng and Feng, Siyuan and Du, Yilun and Xu, Zhenjia and
               Cousineau, Eric and others},
  booktitle= {Proceedings of Robotics: Science and Systems (RSS)},
  year      = {2023},
  month     = {July},
  doi       = {10.15607/RSS.2023.XIX.026},
  publisher = {Robotics: Science and Systems Foundation}
}

@inproceedings{rt2,
  title     = {{RT-2}: Vision-Language-Action Models Transfer Web Knowledge to Robotic Control},
  author    = {Zitkovich, Brianna and Yu, Tianhe and Xu, Sichun and Xu, Peng and
               Xiao, Ted and others},
  booktitle= {Proceedings of the 7th Conference on Robot Learning (CoRL)},
  series    = {Proceedings of Machine Learning Research},
  volume    = {229},
  pages     = {2165--2183},
  year      = {2023},
  publisher = {PMLR}
}

@inproceedings{openvla,
  title     = {{OpenVLA}: An Open-Source Vision-Language-Action Model},
  author    = {Kim, Moo Jin and Pertsch, Karl and Karamcheti, Siddharth and
               Xiao, Ted and Balakrishna, Ashwin and others},
  booktitle= {Proceedings of the 8th Conference on Robot Learning (CoRL)},
  series    = {Proceedings of Machine Learning Research},
  volume    = {270},
  pages     = {2679--2713},
  year      = {2025},
  publisher = {PMLR}
}

@inproceedings{pi0,
  title     = {{$\pi_0$}: A Vision-Language-Action Flow Model for General Robot Control},
  author    = {Black, Kevin and Brown, Noah and Driess, Danny and Esmail, Adnan and
               Equi, Michael Robert and others},
  booktitle= {Proceedings of Robotics: Science and Systems (RSS)},
  year      = {2025},
  month     = {June},
  doi       = {10.15607/RSS.2025.XXI.010},
  publisher = {Robotics: Science and Systems Foundation}
}

@inproceedings{pi0.5,
  title     = {{$\pi_{0.5}$}: A Vision-Language-Action Model with Open-World Generalization},
  author    = {Black, Kevin and Brown, Noah and Darpinian, James and
               Dhabalia, Karan and Driess, Danny and others},
  booktitle= {Proceedings of the 9th Conference on Robot Learning (CoRL)},
  series    = {Proceedings of Machine Learning Research},
  volume    = {305},
  pages     = {17--40},
  year      = {2025},
  publisher = {PMLR}
}

@article{gr00t,
  title   = {{GR00T N1}: An Open Foundation Model for Generalist Humanoid Robots},
  author  = {{NVIDIA} and Bj{\"o}rck, Johan and Casta{\~n}eda, Fernando and
             Cherniadev, Nikita and Da, Xingye and others},
  journal = {arXiv preprint arXiv:2503.14734},
  year    = {2025}
}

@inproceedings{robodreamer,
  title     = {{RoboDreamer}: Learning Compositional World Models for Robot Imagination},
  author    = {Zhou, Siyuan and Du, Yilun and Chen, Jiaben and Li, Yandong and
               Yeung, Dit-Yan and others},
  booktitle= {Proceedings of the 41st International Conference on Machine Learning (ICML)},
  series    = {Proceedings of Machine Learning Research},
  volume    = {235},
  pages     = {61885--61896},
  year      = {2024},
  publisher = {PMLR}
}

@article{lingbotva,
  title   = {Causal World Modeling for Robot Control},
  author  = {Li, Lin and Zhang, Qihang and Luo, Yiming and Yang, Shuai and
             Wang, Ruilin and others},
  journal = {arXiv preprint arXiv:2601.21998},
  year    = {2026}
}

@article{dreamzero,
  title   = {World Action Models Are Zero-Shot Policies},
  author  = {Ye, Seonghyeon and Ge, Yunhao and Zheng, Kaiyuan and Gao, Shenyuan and
             Yu, Sihyun and others},
  journal = {arXiv preprint arXiv:2602.15922},
  year    = {2026}
}

@inproceedings{depthvla,
  title     = {{DepthVLA}: Enhancing Vision-Language-Action Models with Depth-Aware Spatial Reasoning},
  author    = {Yuan, Tianyuan and Liu, Yicheng and Lu, Chenhao and
               Chen, Zhuoguang and Jiang, Tao and others},
  booktitle = {2026 IEEE International Conference on Robotics and Automation (ICRA)},
  year      = {2026},
  publisher= {IEEE}
}

@inproceedings{forcevla,
  title     = {{ForceVLA}: Enhancing {VLA} Models with a Force-Aware {MoE} for Contact-Rich Manipulation},
  author    = {Yu, Jiawen and Liu, Hairuo and Yu, Qiaojun and Ren, Jieji and
               Hao, Ce and others},
  booktitle= {Advances in Neural Information Processing Systems (NeurIPS)},
  volume    = {38},
  year      = {2025},
  doi       = {10.52202/085713-3124},
  pages    = {103609--103639},
  publisher= {Neural Information Processing Systems Foundation, Inc.}
}

@inproceedings{tavla,
  title     = {{TA-VLA}: Elucidating the Design Space of Torque-Aware Vision-Language-Action Models},
  author    = {Zhang, Zongzheng and Xu, Haobo and Yang, Zhuo and Yue, Chenghao and
               Lin, Zehao and others},
  booktitle= {Proceedings of the 9th Conference on Robot Learning (CoRL)},
  series    = {Proceedings of Machine Learning Research},
  volume    = {305},
  pages     = {4019--4037},
  year      = {2025},
  publisher = {PMLR}
}

@article{tactilevla,
  title   = {{Tactile-VLA}: Unlocking Vision-Language-Action Model's Physical Knowledge for Tactile Generalization},
  author  = {Huang, Jialei and Wang, Shuo and Lin, Fanqi and Hu, Yihang and
             Wen, Chuan and others},
  journal = {arXiv preprint arXiv:2507.09160},
  year    = {2025}
}

@inproceedings{safenightvla,
  title     = {{Safe-Night VLA}: Seeing the Unseen via Thermal-Perceptive Vision-Language-Action Models for Safety-Critical Manipulation},
  author    = {Yu, Dian and Zhou, Qingchuan and Huang, Bingkun and
               Khadiv, Majid and Yang, Zewen},
  booktitle = {2026 IEEE/RSJ International Conference on Intelligent Robots and Systems (IROS)},
  year      = {2026},
  publisher= {IEEE}
}

@inproceedings{omnivla,
  title     = {{OmniVLA}: Physically-Grounded Multimodal {VLA} with Unified Multi-Sensor Perception for Robotic Manipulation},
  author    = {Guo, Heyu and Wang, Shanmu and Ma, Ruichun and Jiang, Shiqi and
               Ghasempour, Yasaman and others},
  booktitle = {2026 IEEE International Conference on Robotics and Automation (ICRA)},
  year      = {2026},
  publisher= {IEEE}
}

@article{musevla,
  title   = {{MuseVLA}: An Adaptive Multimodal Sensing Vision-Language-Action Model for Robotic Manipulation},
  author  = {Liu, Xingyuming and Ma, Ruichun and Guo, Heyu and Li, Qixiu and
             Yang, Qingwen and others},
  journal = {arXiv preprint arXiv:2606.17598},
  year    = {2026}
}

@article{cmx,
  title   = {{CMX}: Cross-Modal Fusion for {RGB-X} Semantic Segmentation with Transformers},
  author  = {Zhang, Jiaming and Liu, Huayao and Yang, Kailun and
             Hu, Xinxin and Liu, Ruiping and others},
  journal = {IEEE Transactions on Intelligent Transportation Systems},
  volume  = {24},
  number  = {12},
  pages   = {14679--14694},
  year    = {2023},
  doi     = {10.1109/TITS.2023.3300537},
  publisher= {IEEE}
}

@inproceedings{anythermal,
  title     = {{AnyThermal}: Towards Learning Universal Representations for Thermal Perception},
  author    = {Maheshwari, Parv and Karhade, Jay and Chawla, Yogesh and
               Adu, Isaiah and Heisen, Florian and others},
  booktitle = {2026 IEEE International Conference on Robotics and Automation (ICRA)},
  year      = {2026},
  publisher= {IEEE}
}

@inproceedings{irgpt,
  title     = {{IRGPT}: Understanding Real-World Infrared Image with Bi-Cross-Modal Curriculum on Large-Scale Benchmark},
  author    = {Cao, Zhe and Zhang, Jin and Zhang, Ruiheng},
  booktitle= {Proceedings of the IEEE/CVF International Conference on Computer Vision (ICCV)},
  pages     = {166--176},
  year      = {2025},
  publisher= {IEEE}
}

@article{thermovl,
  title   = {{Thermo-VL}: Extending Vision-Language Models to Thermal Infrared Perception},
  author  = {Thushara, Rusiru and Ranasinghe, Yasiru and
             Paranjape, Jay and Patel, Vishal M.},
  journal = {arXiv preprint arXiv:2605.21882},
  year    = {2026}
}

@article{thermoact,
  title   = {{ThermoAct}: Thermal-Aware Vision-Language-Action Models for Robotic Perception and Decision-Making},
  author  = {Son, Young-Chae and Ko, Dae-Kwan and Choi, Yoon-Ji and
             Lim, Soo-Chul},
  journal = {IEEE Robotics and Automation Letters},
  volume  = {11},
  number  = {5},
  pages   = {6106--6113},
  year    = {2026},
  month   = {May},
  doi     = {10.1109/LRA.2026.3678130},
  publisher= {IEEE}
}

@misc{gemini,
  title        = {Gemini 2.0 Flash},
  author       = {{Google DeepMind}},
  year         = {2025},
  howpublished = {Model Card},
  publisher= {Google DeepMind}
}

@inproceedings{vqbet,
  title     = {Behavior Generation with Latent Actions},
  author    = {Lee, Seungjae and Wang, Yibin and Etukuru, Haritheja and
               Kim, H. Jin and Shafiullah, Nur Muhammad Mahi and others},
  booktitle= {Proceedings of the 41st International Conference on Machine Learning (ICML)},
  series    = {Proceedings of Machine Learning Research},
  volume    = {235},
  pages     = {26991--27008},
  year      = {2024},
  publisher = {PMLR}
}

@article{pi0.7,
  title   = {{$\pi_{0.7}$}: A Steerable Generalist Robotic Foundation Model with Emergent Capabilities},
  author  = {{Physical Intelligence} and Ai, Bo and Amin, Ali and
             Aniceto, Raichelle and Balakrishna, Ashwin and others},
  journal = {arXiv preprint arXiv:2604.15483},
  year    = {2026}
}

@inproceedings{gr1,
  title     = {Unleashing Large-Scale Video Generative Pre-Training for Visual Robot Manipulation},
  author    = {Wu, Hongtao and Jing, Ya and Cheang, Chilam and
               Chen, Guangzeng and Xu, Jiafeng and others},
  booktitle= {International Conference on Learning Representations (ICLR)},
  year      = {2024}
}

@article{gr2,
  title   = {{GR-2}: A Generative Video-Language-Action Model with Web-Scale Knowledge for Robot Manipulation},
  author  = {Cheang, Chi-Lam and Chen, Guangzeng and Jing, Ya and
             Kong, Tao and Li, Hang and others},
  journal = {arXiv preprint arXiv:2410.06158},
  year    = {2024}
}

@article{vtla,
  title   = {{VTLA}: Vision-Tactile-Language-Action Model with Preference Learning for Insertion Manipulation},
  author  = {Zhang, Chaofan and Hao, Peng and Cao, Xiaoge and
             Hao, Xiaoshuai and Cui, Shaowei and others},
  journal = {Biomimetic Intelligence and Robotics},
  pages   = {100333},
  year    = {2026},
  doi     = {10.1016/j.birob.2026.100333},
  publisher= {Elsevier}
}

@inproceedings{tacvla,
  title     = {{TacVLA}: Contact-Aware Tactile Fusion for Robust Vision-Language-Action Manipulation},
  author    = {Zhang, Kaidi and Zhang, Heng and Xu, Zhengtong and
               Zhang, Zhiyuan and Prince, Md Rakibul Islam and others},
  booktitle = {2026 IEEE/RSJ International Conference on Intelligent Robots and Systems (IROS)},
  year      = {2026},
  publisher= {IEEE}
}

@inproceedings{audiovla,
  title     = {{Audio-VLA}: Adding Contact Audio Perception to Vision-Language-Action Model for Robotic Manipulation},
  author    = {Wei, Xiangyi and Zhang, Haotian and Cao, Xinyi and
               Xie, Siyu and Ge, Weifeng and others},
  booktitle = {2026 IEEE International Conference on Robotics and Automation (ICRA)},
  year      = {2026},
  publisher= {IEEE}
}

@inproceedings{evla,
  title     = {{E-VLA}: Event-Augmented Vision-Language-Action Model for Dark and Blurred Scenes},
  author    = {Zhai, Jiajun and Shi, Hao and Guo, Shangwei and
               Yang, Kailun and Wang, Kaiwei},
  booktitle = {European Conference on Computer Vision (ECCV)},
  year      = {2026}
}

@article{raccm,
  title   = {Region-Adaptive and Context-Complementary Cross Modulation for {RGB-T} Semantic Segmentation},
  author   = {Peng, Fengguang and Ding, Zihan and Chen, Ziming and Wang, Gang and Hui, Tianrui and others},
  journal = {Pattern Recognition},
  volume  = {147},
  pages   = {110092},
  year    = {2024},
  doi     = {10.1016/j.patcog.2023.110092},
  publisher= {Elsevier}
}

@inproceedings{freqfusion,
  title     = {Frequency-Guided Fusion for {RGB}-Thermal Semantic Segmentation},
  author    = {Canitez, Ismail Emre and Erkent, Ozgur},
  booktitle= {Proceedings of the IEEE/CVF Conference on Computer Vision and Pattern Recognition Workshops (CVPR Workshops)},
  pages     = {7018--7026},
  year      = {2026},
  publisher= {IEEE}
}

@article{eym,
  title   = {The Approximation of One Matrix by Another of Lower Rank},
  author  = {Eckart, Carl and Young, Gale},
  journal = {Psychometrika},
  volume  = {1},
  number  = {3},
  pages   = {211--218},
  year    = {1936},
  doi     = {10.1007/BF02288367},
  publisher= {Springer}
}

@inproceedings{resnet,
  title     = {Deep Residual Learning for Image Recognition},
  author    = {He, Kaiming and Zhang, Xiangyu and Ren, Shaoqing and Sun, Jian},
  booktitle= {Proceedings of the IEEE Conference on Computer Vision and Pattern Recognition (CVPR)},
  pages     = {770--778},
  year      = {2016},
  doi       = {10.1109/CVPR.2016.90},
  publisher= {IEEE}
}

@inproceedings{minima,
  author    = {Jiangwei Ren and Xingyu Jiang and Zizhuo Li and Dingkang Liang and Xin Zhou and others},
  title     = {MINIMA: Modality Invariant Image Matching},
  booktitle = {Proceedings of the IEEE/CVF Conference on Computer Vision and Pattern Recognition (CVPR)},
  pages     = {23059--23068},
  year      = {2025},
  publisher= {IEEE},
  doi      = {10.1109/CVPR52734.2025.02147}
}

@inproceedings{superpoint,
  author    = {Daniel DeTone and Tomasz Malisiewicz and Andrew Rabinovich},
  title     = {SuperPoint: Self-Supervised Interest Point Detection and Description},
  booktitle = {Proceedings of the IEEE Conference on Computer Vision and Pattern Recognition Workshops (CVPR Workshops)},
  pages     = {224--236},
  year      = {2018},
  publisher= {IEEE}
}

@inproceedings{lightglue,
  author    = {Philipp Lindenberger and Paul-Edouard Sarlin and Marc Pollefeys},
  title     = {LightGlue: Local Feature Matching at Light Speed},
  booktitle = {Proceedings of the IEEE/CVF International Conference on Computer Vision (ICCV)},
  pages     = {17627--17638},
  year      = {2023},
  publisher= {IEEE},
  doi      = {10.1109/ICCV51070.2023.01616}
}

@article{eckart1936approximation,
  author  = {Carl Eckart and Gale Young},
  title   = {The Approximation of One Matrix by Another of Lower Rank},
  journal = {Psychometrika},
  volume  = {1},
  number  = {3},
  pages   = {211--218},
  year    = {1936},
  doi     = {10.1007/BF02288367},
  publisher= {Springer}
}

@article{mirsky1960symmetric,
  author  = {L. Mirsky},
  title   = {Symmetric Gauge Functions and Unitarily Invariant Norms},
  journal = {The Quarterly Journal of Mathematics},
  volume  = {11},
  number  = {1},
  pages   = {50--59},
  year    = {1960},
  doi     = {10.1093/qmath/11.1.50},
  publisher= {Oxford University Press}
}

@inproceedings{robotseg,
  author    = {Haiyang Mei and Qiming Huang and Hai Ci and Mike Zheng Shou},
  title     = {RobotSeg: A Model and Dataset for Segmenting Robots in Image and Video},
  booktitle = {Proceedings of the IEEE/CVF Conference on Computer Vision and Pattern Recognition (CVPR)},
  pages     = {14346--14356},
  year      = {2026},
  publisher= {IEEE}
}

@inproceedings{cutie,
  author    = {Ho Kei Cheng and Seoung Wug Oh and Brian Price and Joon-Young Lee and Alexander Schwing},
  title     = {Putting the Object Back into Video Object Segmentation},
  booktitle = {Proceedings of the IEEE/CVF Conference on Computer Vision and Pattern Recognition (CVPR)},
  pages     = {3151--3161},
  year      = {2024},
  publisher= {IEEE},
  doi      = {10.1109/CVPR52733.2024.00304}
}

@article{matchanything,
  title   = {{MatchAnything}: Universal Cross-Modality Image Matching with Large-Scale Pre-Training},
  author  = {He, Xingyi and Yu, Hao and Peng, Sida and Tan, Dongli and Shen, Zehong and others},
  journal = {arXiv preprint arXiv:2501.07556},
  year    = {2025}
}

@inproceedings{romav2,
  title     = {{RoMa v2}: Harder Better Faster Denser Feature Matching},
  author    = {Edstedt, Johan and Nordstr{\"o}m, David and Zhang, Yushan and B{\"o}kman, Georg and Astermark, Jonathan and others},
  booktitle = {European Conference on Computer Vision (ECCV)},
  year      = {2026}
}

@inproceedings{xoftr,
  title     = {{XoFTR}: Cross-Modal Feature Matching Transformer},
  author    = {Tuzcuo{\u{g}}lu, {\"O}nder and K{\"o}ksal, Aybora and Sofu, Bu{\u{g}}ra and Kalkan, Sinan and Alatan, A. Aydin},
  booktitle = {2024 IEEE/CVF Conference on Computer Vision and Pattern Recognition Workshops (CVPR Workshops)},
  pages     = {4275--4286},
  year      = {2024},
  publisher = {IEEE},
  doi       = {10.1109/CVPRW63382.2024.00431}
}

@inproceedings{dkm,
  title     = {{DKM}: Dense Kernelized Feature Matching for Geometry Estimation},
  author    = {Edstedt, Johan and Athanasiadis, Ioannis and Wadenb{\"a}ck, M{\aa}rten and Felsberg, Michael},
  booktitle = {2023 IEEE/CVF Conference on Computer Vision and Pattern Recognition (CVPR)},
  pages     = {17765--17775},
  year      = {2023},
  publisher = {IEEE}
}

@inproceedings{roma,
  title     = {{RoMa}: Robust Dense Feature Matching},
  author    = {Edstedt, Johan and Sun, Qiyu and B{\"o}kman, Georg and Wadenb{\"a}ck, M{\aa}rten and Felsberg, Michael},
  booktitle = {2024 IEEE/CVF Conference on Computer Vision and Pattern Recognition (CVPR)},
  pages     = {19790--19800},
  year      = {2024},
  publisher = {IEEE},
  doi       = {10.1109/CVPR52733.2024.01871}
}

\end{document}


\maketitle

\section{Symbol Notation}
For clarity and consistency, Table~\ref{tab:notation} summarises the main
notation used throughout this work, including observations, visual and
attribute representations, model components, gating variables, robot states and actions, and flow-matching variables.
\begin{longtblr}[
  caption = {\textbf{Notation.}},
  label = {tab:notation},
]{
  width = \columnwidth,
  colspec = {Q[l,wd=0.20\columnwidth] Q[l,wd=0.72\columnwidth]},
  rowhead = 1,
  row{1} = {font=\bfseries},
  hline{1,2,Z} = {0.8pt},
}

Symbol & Description \\

$x$
& Original RGB visual representation \\

$a$
& Attribute observation  \\

$l$
& Language instruction or condition \\

$f(\cdot)$
& Function that outputs the additive residual \\

$y$
& Attribute-modulated RGB representation \\

$x_r$
& RGB observations \\

$x_r^i$
& RGB observation from camera $i$ \\

$x_t$
& Thermal observation \\

$x_{\mathrm{cold}}$
& Cold attribute view \\

$x_{\mathrm{hot}}$
& Hot attribute view \\

$x_{\mathrm{col}}^j$
& Colour attribute view for colour $j$ \\

$Z_{\mathrm{rgb}}^i$
& RGB tokens from camera $i$ \\

$Z_{\mathrm{rgb}}^{\mathrm{head}}$
& Head-camera RGB tokens \\

$Z_{\mathrm{rgb}}^{\mathrm{left}}$
& Left wrist-camera RGB tokens \\

$Z_{\mathrm{rgb}}^{\mathrm{right}}$
& Right wrist-camera RGB tokens \\

$Z_{\mathrm{cold}}$
& Cold attribute tokens \\

$Z_{\mathrm{hot}}$
& Hot attribute tokens \\

$Z_{\mathrm{rgb}}^{\mathrm{cold}}$
& Cold-conditioned RGB representation \\

$Z_{\mathrm{rgb}}^{\mathrm{hot}}$
& Hot-conditioned RGB representation \\

$Z_{\mathrm{col}}^j$
& Colour attribute tokens for colour $j$ \\

$Z_{\mathrm{rgb}}^j$
& Colour-conditioned RGB representation for colour $j$ \\

$Z_{\mathrm{text}}$
& Joint textual representation of language and robot state \\

$Z_T$
& Temperature-conditioned multimodal prefix \\

$E_r$
& RGB visual encoder \\

$E_t$
& Shared thermal visual encoder \\

$E_{\mathrm{col}}$
& Shared colour visual encoder \\

$P_r$
& Visual projector \\

$P_{\mathrm{col}}$
& Colour visual projector \\

$O_c$
& Cross-attention output of the cold branch \\

$O_h$
& Cross-attention output of the hot branch \\

$u_t$
& Gating feature \\

$\alpha,\beta$
& Dynamic weights for cold and hot attributes \\

$\alpha',\beta',\gamma$
& Dynamic weights for red, green, and blue attributes \\

$\operatorname{LN}$
& Layer normalization \\

$\operatorname{MLP}$
& Gated multilayer perceptron \\

$s_t$
& Robot state at time step $t$ \\

$a_t$
& Robot action at time step $t$ \\

$A_t$
& Action chunk starting at time step $t$ \\

$H$
& Action horizon \\

$d_a$
& Single-arm action dimension \\

$\pi_\theta$
& Parameterised action policy \\

$\theta$
& Learnable policy parameters \\

$i$
& Camera index, $i\in\{\mathrm{head},\mathrm{left},\mathrm{right}\}$ \\

$j$
& Colour index, $j\in\{\mathrm{red},\mathrm{green},\mathrm{blue}\}$ \\

$B$
& Batch size \\

$N$
& Number of visual tokens \\

$D$
& Token dimensionality \\

$b$
& Background reference thermal response \\

$\mu$
& Fixed background grayscale intensity \\

$k$
& Background-calibrated thermal observation \\

$I_{\max}$
& Upper bound of the grayscale range \\

$\epsilon$
& Gaussian noise, $\epsilon\sim\mathcal{N}(0,I)$ \\

$I$
& Identity matrix \\

$\tau$
& Flow-matching time step \\

$x_\tau$
& Interpolated action variable at flow time $\tau$ \\

\end{longtblr}
\clearpage

\section{Hardware Details}

\begin{figure}[t]
\centering
\includegraphics[width=\linewidth]{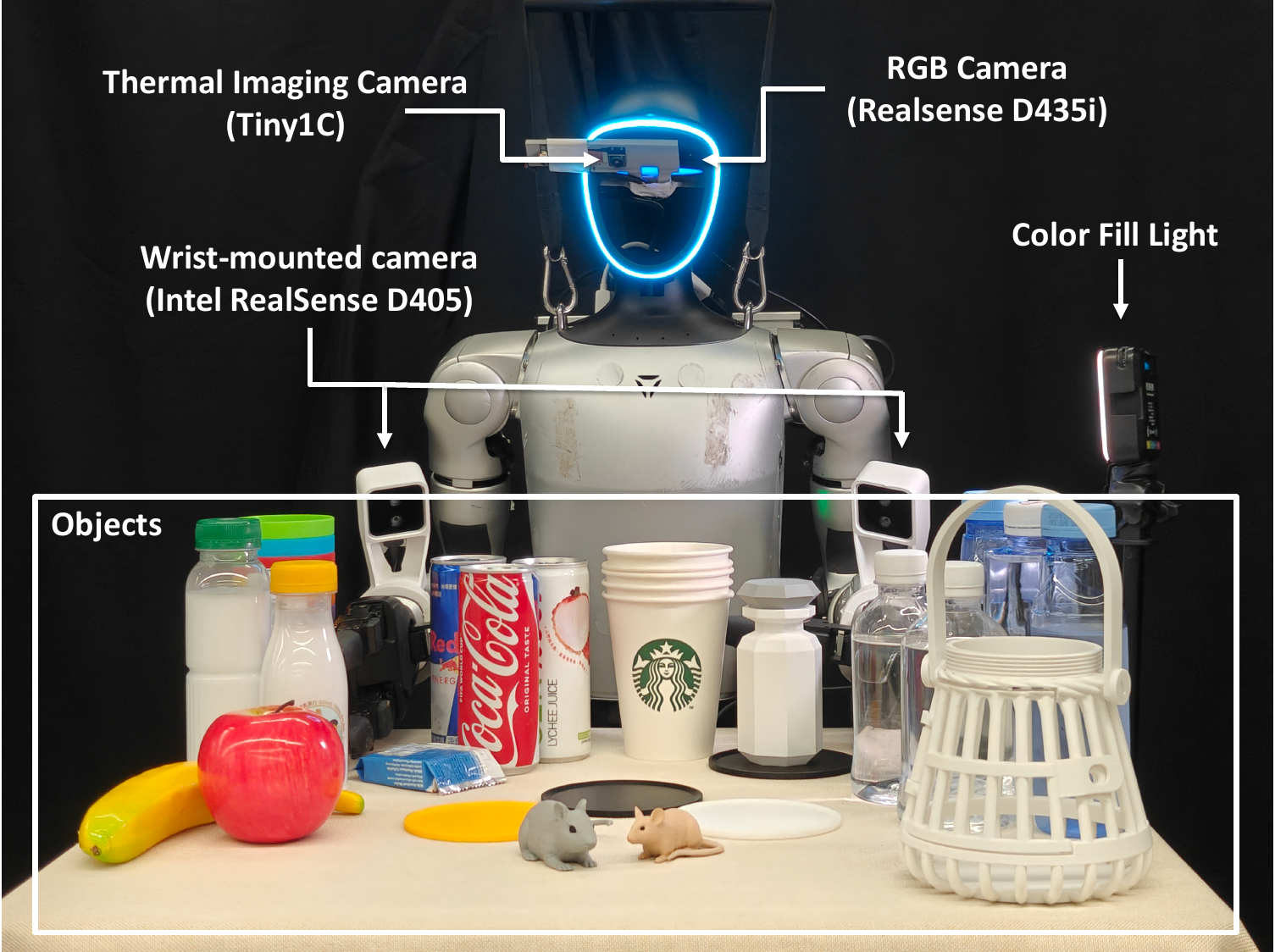}
\caption{\textbf{Equipment}: Unitree G1 EDU platform, sensing setup, and manipulation objects across 7 tasks.}

\label{g1}
\end{figure}

Experiments are conducted on a Unitree G1 EDU humanoid robot equipped with Unitree Dex3-1 three-finger dexterous hands, with only the two arms and hands being controlled. Visual observations are captured by a head-mounted Intel RealSense D435i RGB camera (640$\times$480, 30 Hz), two wrist-mounted Intel RealSense D405 cameras, and a Tiny1-C thermal camera (256$\times$192, 3.2 mm, 25 Hz) mounted to the right of the D435i. Model training is performed on an NVIDIA A800 80GB GPU, while policy inference runs on an NVIDIA GeForce RTX 5090 48GB GPU. Both systems use Ubuntu 22.04.
As shown in Fig.~\ref{g1}, the RGB and thermal cameras are rigidly mounted on the robot head, while the two wrist cameras are installed on the left and right wrists following Unitree's official mounting configuration. The figure also presents the complete set of manipulation objects used throughout our experiments, including objects used for task data collection and additional objects introduced for unseen-object generalisation evaluation. Several objects are fabricated via 3D printing using publicly available models from MakerWorld.
\section{Training Details}
\subsection{Training Budget Selection}
\begin{figure}[h]
\centering
\includegraphics[width=\linewidth]{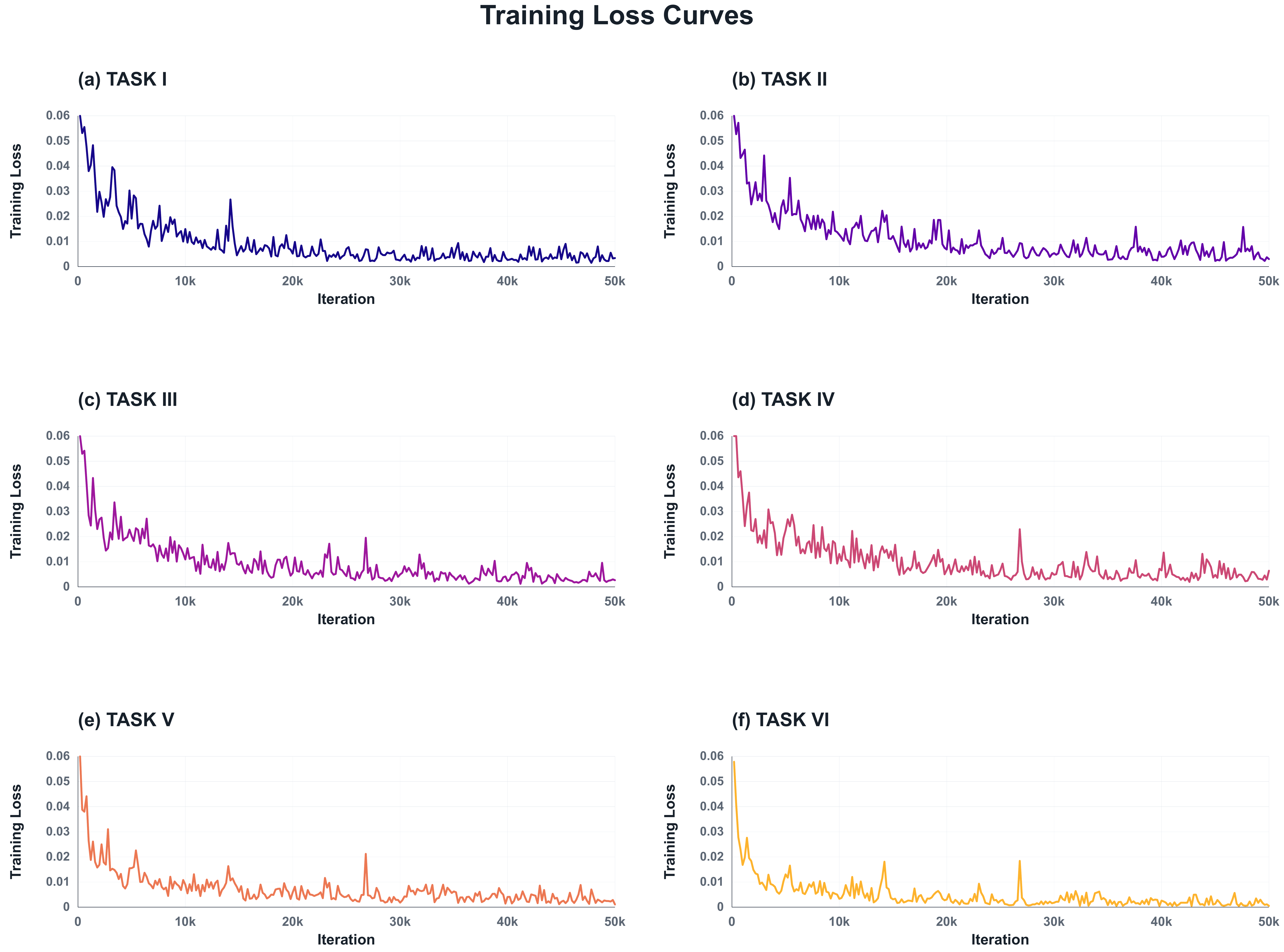}
\caption{\textbf{Training loss curves.}
All tasks converge to stable loss values within the 40K-step fine-tuning budget used in our experiments.}
\label{loss}
\end{figure}

Fig.~\ref{loss} shows the training loss curves for the six tasks. Across all tasks, the loss decreases rapidly during the early stage of training and gradually converges after approximately 20K--30K iterations. By 40K iterations, the loss curves have reached consistently low and stable levels, while further training yields only marginal reductions with continued stochastic fluctuations. We therefore use 40K iterations as the training budget and select the corresponding checkpoint for evaluation. For reproducibility, the random seed is fixed to 1000 for all experiments.


\section{Dataset}
\subsection{XR Teleoperation}
Datasets are collected following the XR teleoperation pipeline provided in Unitree Robotics' official XR teleoperate repository. We use a PICO XR system and develop a semi-automated collection procedure, where each recorded episode includes an automatic transition from the predefined initial pose to the operator's current pose, teleoperated task execution, and an automatic return to the initial pose after task completion. In addition to the original robot states and RGB observations, thermal observations are synchronously recorded and integrated into an extended LeRobot v3 data format. This unified multimodal dataset is used for all VLA training and evaluation in this work.

\subsection{Thermal Objects Elimination}

In Task~I and Task~III, the target objects should not provide task-relevant temperature cues, while the thermal radiation from the robot hands should still be preserved. To reuse the thermal data collected in Task~II and Task~IV, we remove the thermal responses of target objects and retain only the robot-hand heat signatures. We explore two segmentation approaches, RobotSeg and Cutie.

\begin{figure}[t]
\centering
\includegraphics[width=\linewidth]{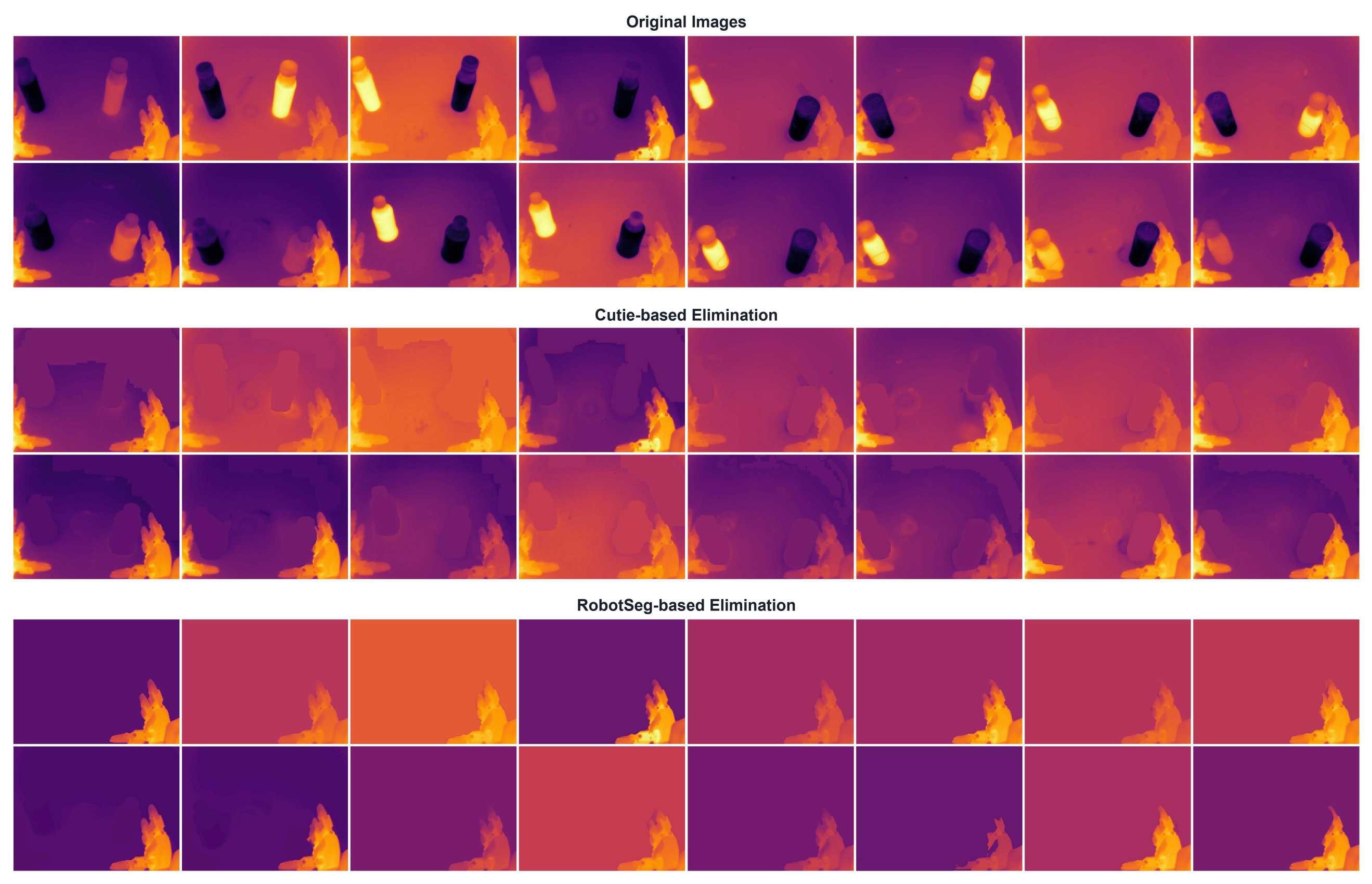}
\caption{\textbf{Thermal-object elimination for constructing RGB-only observations.}
RobotSeg directly extracts the robot regions but may lose the hand under occlusion, whereas Cutie tracks and removes temperature-bearing target objects more consistently while preserving the thermal responses of the robot hands.}

\label{elimination}
\end{figure}

$ $
\noindent{\normalfont{\textbf{RobotSeg.}}} RobotSeg~\cite{robotseg} is a robot-specific segmentation model built upon SAM~2, combining structure-enhanced temporal memory with an automatic robot prompt generator. We use it to segment the robot regions in thermal sequences, retain the arm and hand signals, and replace the remaining regions with the background mean value. While RobotSeg can reliably localise a visible arm, it may fail when the hand becomes occluded, as shown in Fig.~\ref{elimination}.

$ $
\noindent{\normalfont{\textbf{Cutie.}}} Cutie~\cite{cutie} is a video object segmentation model based on object-level memory and query-based feature interaction. We manually annotate the temperature-bearing target objects in a few key frames and propagate their masks throughout the sequence using Cutie. The segmented target regions are then removed, followed by thresholding to suppress residual low-temperature responses. Compared with RobotSeg, Cutie provides more stable segmentation and produces cleaner thermal observations consistent with the RGB-only settings of Task~I and Task~III. Qualitative results are shown in Fig.~\ref{elimination}.


\subsection{Dataset Details}

\begin{table*}[!p]
\centering

\caption{\textbf{Dataset details.}
Language instructions and number of collected episodes for each task. For Tasks~I and II, $x+y$ denotes episodes collected with the two bottle types, where $x$ and $y$ correspond to the shorter and blue bottles, respectively.}
\label{tab}

\renewcommand{\arraystretch}{1.15}
\begin{tabular}{llc}
\toprule
\textbf{Task Name} & \textbf{Language Instructions} & \textbf{Episode} \\
\midrule

\multicolumn{3}{l}{\textit{\textbf{Task 1: Position-guided bottle selection}}} \\ 
 & Pick up the bottle on the left and place it on the black circular coaster. & 39 + 42  \\
 & Pick up the bottle on the right and place it on the black circular coaster. & 46 + 29  \\
\midrule

\multicolumn{3}{l}{\textit{\textbf{Task 2: Temperature-guided bottle selection}}} \\ 
 & Pick up the cold bottle and place it on the black circular coaster. & 152 + 151 \\
\midrule

\multicolumn{3}{l}{\textit{\textbf{Task 3: Identity-guided object selection}}} \\ 
 & Pick up the milk and place it on the white circular coaster. & 48 \\
 & Pick up the milk and place it on the yellow circular coaster. & 56  \\ 
 & Pick up the coke and place it on the white circular coaster. & 71  \\
 & Pick up the coke and place it on the yellow circular coaster. & 61  \\ 
\midrule

\multicolumn{3}{l}{\textit{\textbf{Task 4: Temperature-guided object selection}}} \\ 
 & Pick up the hot object and place it on the yellow circular coaster. & 89 \\
 & Pick up the hot object and place it on the white circular coaster. & 84  \\ 
 & Pick up the cold object and place it on the yellow circular coaster. & 83  \\
 & Pick up the cold object and place it on the white circular coaster. & 84  \\ 
\midrule

\multicolumn{3}{l}{\textit{\textbf{Task 5: Temperature-aware coffee service}}} \\ 
 & Pick up the sugar container and add sugar to the coffee with less ice. & 87 \\ 
 & Pick up the sugar container and add sugar to the coffee with extra ice. & 86 \\
\midrule
           
\multicolumn{3}{l}{\textit{\textbf{Task 6: Abnormal-animal cage removal}}} \\ 
 & Select the cage with the dead mouse and place it in the hazard-marked box. & 99  \\
\midrule

\multicolumn{3}{l}{\textit{\textbf{Task 7: Colour-guided cube selection (Simulation)}}} \\ 
 & Put the red cube on the plate & 50  \\
 & Put the green cube on the plate & 50  \\
 & Put the blue cube on the plate & 50  \\
\midrule

\multicolumn{3}{l}{\textit{\textbf{Task 8: Colour-guided cup selection}}} \\ 
 & Select the red cup and place it on the black coaster & 73  \\
 & Select the blue cup and place it on the black coaster & 51  \\

\bottomrule
\end{tabular}

\vspace{5mm}

\caption{\textbf{Implementation configuration of the ThermoRes Encoder.}
The table summarises the backbone and input settings, thermal attribute decomposition, visual encoding and gating, and action-generation configuration.}
\label{tab:dualbranch-encoder}

\small
\setlength{\tabcolsep}{6pt}
\renewcommand{\arraystretch}{1.10}

\begin{tabular}{p{0.26\textwidth} p{0.37\textwidth} p{0.28\textwidth}}
\toprule
\textbf{Component} & \textbf{Configuration} & \textbf{Value} \\
\midrule

\multicolumn{3}{l}{\textbf{Backbone and Input}} \\[-2pt]
\addlinespace[2pt]

Backbone
& Vision--language backbone / action expert
& $2$B / $300$M parameters \\

Input geometry
& Image resolution / visual tokens / token dimension
& $224{\times}224$ / $256$ / $2048$ \\

Sequence configuration
& Maximum textual length / action horizon
& $200$ / $50$ \\

\midrule
\multicolumn{3}{l}{\textbf{Thermal Attribute Separation}} \\[-2pt]
\addlinespace[2pt]

Background estimation
& Spatial sampling regions
& $3\times64{\times}64$ upper-boundary regions \\

Background alignment
& Reference grayscale intensity
& $\mu=80$ on $[0,255]$ \\

Outlier filtering
& Robust threshold / relative threshold
& $12$ / $2.0$ \\

\midrule
\multicolumn{3}{l}{\textbf{Visual Encoding and Gating}} \\[-2pt]
\addlinespace[2pt]

Visual encoders
& RGB encoder / thermal encoder
& Independent / shared across 2 branches \\

Visual projection
& Projector across visual streams
& Shared $P_r$ \\

Cross-attention
& Textual query / thermal key--value
& $Q=Z_{\mathrm{text}},\;
K=V\in\{Z_{\mathrm{cold}},Z_{\mathrm{hot}}\}$ \\

Attention configuration
& Number of heads / head dimension
& $8$ / $256$ \\

Gate network
& MLP dimensions
& $10240\!\rightarrow\!512\!\rightarrow\!256\!\rightarrow\!2$ \\

Gate activation
& Hidden activation / dropout
& GELU / $0$ \\

Gate initialization
& Output weight / bias
& $\mathcal{N}(0,0.02)$ / $0$ \\

\midrule
\multicolumn{3}{l}{\textbf{Output and Action Generation}} \\[-2pt]
\addlinespace[2pt]

Dynamic gating
& Cold / hot branch weights
& $[\alpha,\beta]
=\,\mathrm{softmax}(\mathrm{MLP}(\mathrm{LN}(u_t)))$ \\

Action generation
& Action embedding dimension / integration steps
& $1024$ / $10$ \\

\bottomrule
\end{tabular}

\end{table*}

Our dataset is designed to evaluate both conventional RGB manipulation capability and progressively more complex thermal reasoning. Tasks~I and III are RGB-only tasks based on spatial position and object identity, respectively, while Tasks~II, IV, V, and VI progressively increase the thermal reasoning difficulty: from binary temperature-based selection, to target selection among multiple objects, fine-grained temperature discrimination in a drink-service scenario, and abnormality reasoning in a laboratory scenario. Tasks~V and VI further move beyond basic pick-and-place settings toward more practical and long-horizon manipulation.

Tasks~VII and VIII further examine whether the proposed residual formulation can generalise beyond thermal attributes by using colour as the task-relevant cue. Task~VII is conducted in simulation because existing robotic simulators do not provide sufficiently realistic thermal observations, while Task~VIII provides the corresponding real-world evaluation. Table~\ref{tab} summarises the language instructions and number of collected episodes for all tasks.

\subsection{Training Parameters}
All $\pi_{0.5}$-based models are initialised from the same pretrained $\pi_{0.5}$ checkpoint and fine-tuned under identical settings with AdamW for 40K steps (batch size 8, learning rate $2.5 \times 10^{-5}$, 100-step warm-up, and cosine decay). Other baselines follow their corresponding architectures and training configurations as specified in their original papers, while using the same task data and evaluation protocol; further implementation details are provided in Table~\ref{tab:dualbranch-encoder}.

\section{Thermo-VLA with RGB--Thermal Alignment}
\subsection{Motivation}
The current implementation of Thermo-VLA uses rigidly mounted RGB and thermal cameras, such that their relative pose remains fixed during both data collection and inference. Under this setting, it is first important to examine whether explicit RGB--thermal matching is necessary, or whether the learned multimodal representation can already tolerate the fixed spatial offset between the two sensing streams. Meanwhile, in more general robotic deployments, the thermal camera may be placed at different positions or viewpoints, leading to larger cross-modal misalignment. We therefore investigate explicit cross-modal matching both to assess its necessity under the default fixed-viewpoint setting and to explore a potential extension that relaxes Thermo-VLA's dependence on a fixed RGB--thermal configuration without modifying its core architecture.

\begin{figure}[t]
\centering
\includegraphics[width=\linewidth]{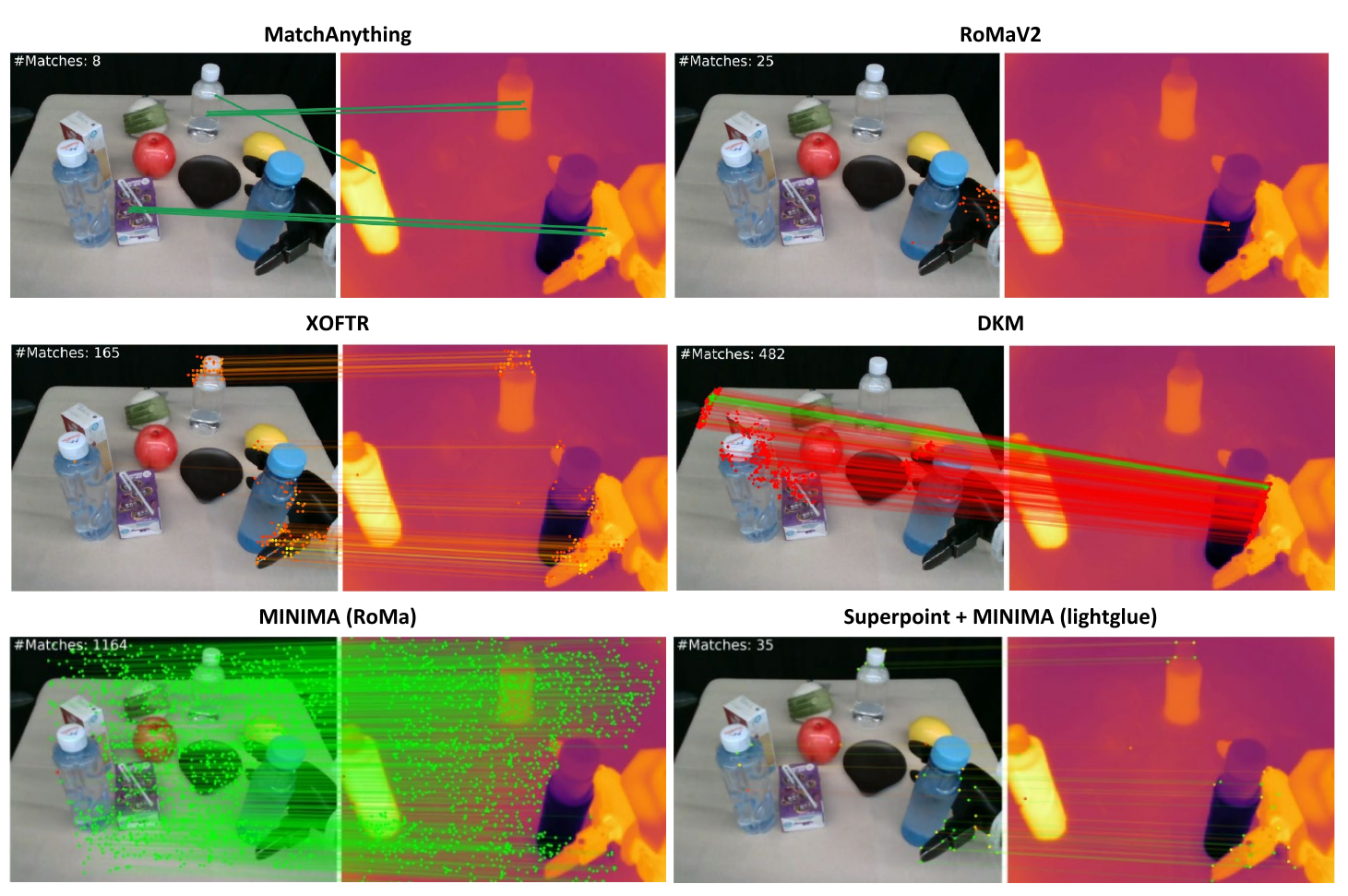}
\caption{\textbf{Comparison of RGB--thermal cross-modal matching methods.}
We compare MatchAnything, RoMaV2, XoFTR, DKM, MINIMA (RoMa), and SuperPoint + MINIMA (LightGlue). Among the compared methods, SuperPoint + MINIMA (LightGlue) provides the most stable correspondences in our setting and is therefore used in our explicit-alignment extension.}
\label{minima}
\end{figure}

\subsection{Matching Algorithm}

To identify a suitable RGB--thermal matching method, we compare MatchAnything~\cite{matchanything}, RoMaV2~\cite{romav2}, XoFTR~\cite{xoftr}, DKM~\cite{dkm}, MINIMA (RoMa)~\cite{minima,roma}, and SuperPoint + MINIMA (LightGlue)~\cite{superpoint,minima,lightglue}, as shown in Fig.~\ref{minima}. The MINIMA-based methods yield more reliable cross-modal correspondences, with SuperPoint + MINIMA (LightGlue) providing the most stable results in our setting. We therefore adopt this configuration.

SuperPoint~\cite{superpoint} extracts sparse local features from RGB and thermal images, while the MINIMA-adapted LightGlue~\cite{lightglue,minima} establishes correspondences for spatial pre-alignment before thermal preprocessing. Under the same tasks and viewpoints as the main experiments, explicit matching achieves comparable success rates to standard Thermo-VLA but introduces additional inference latency. This suggests that explicit alignment is unnecessary under the default fixed-viewpoint setting, while providing a practical extension when larger RGB--thermal misalignment occurs.

\begin{figure*}[t]
\centering
\includegraphics[width=\linewidth]{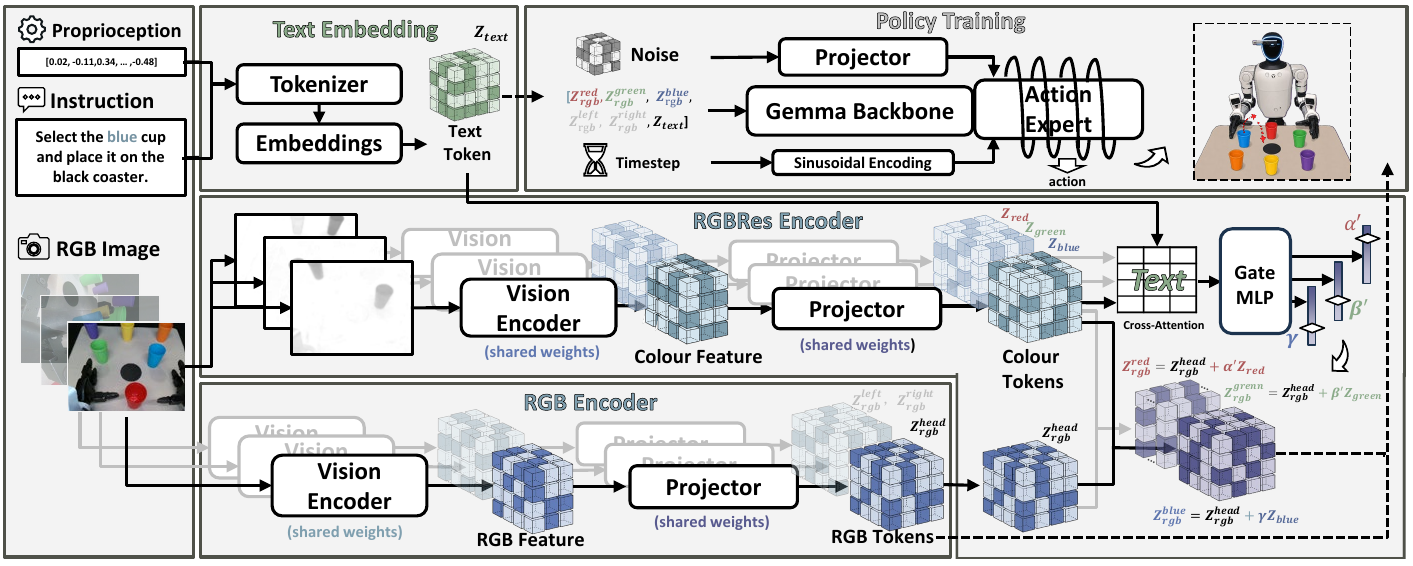}
\caption{\textbf{Overview of $\pi_{0.5}$ with the RGBRes Encoder.}
Attribute-specific RGB views are encoded as residual representations and injected into the original RGB representation through language-conditioned gating, extending the residual formulation beyond thermal attributes.}
\label{pipeline2}
\end{figure*}

\section{Overview of pi0.5 with RGBRes Encoder}
As shown in Fig.\ref{pipeline2}, we extend $\pi_{0.5}$ with an additional RGBRes Encoder to introduce attribute-conditioned residual modulation into the RGB visual representation.

\section{Low-Rank Analysis of Residual Modulation}
\subsection{Preliminaries}
We first recall the classical Eckart--Young--Mirsky theorem for optimal low-rank approximation~\cite{eckart1936approximation,mirsky1960symmetric}.

$ $

\noindent\textbf{Theorem 1 (Eckart--Young--Mirsky).}
Let $M\in\mathbb{R}^{m\times d}$ be a matrix of rank $r$, with singular value decomposition
\begin{equation*}
M=U\Sigma V^{\top}=\sum_{j=1}^{r}\sigma_j u_jv_j^{\top},
\end{equation*}
where
\begin{equation*}
\sigma_1\geq\sigma_2\geq\cdots\geq\sigma_r>0.
\end{equation*}

For any integer $1\leq k<r$, define the rank-$k$ truncated singular value decomposition as
\begin{equation*}
M_k=\sum_{j=1}^{k}\sigma_j u_jv_j^{\top}.
\end{equation*}

Then $M_k$ is a best rank-$k$ approximation of $M$ under the Frobenius norm, i.e.,
\begin{equation*}
M_k\in\underset{\operatorname{rank}(A)\leq k}{\arg\min}\|M-A\|_F,
\end{equation*}
and the minimum approximation error is
\begin{equation*}
\min_{\operatorname{rank}(A)\leq k}\|M-A\|_F=\|M-M_k\|_F=
\left(\sum_{j=k+1}^{r}\sigma_j^2\right)^{1/2}.
\end{equation*}
Equivalently,
\begin{equation*}
\min_{\operatorname{rank}(A)\leq k}\|M-A\|_F^2=\sum_{j=k+1}^{r}\sigma_j^2.
\end{equation*}
If $\sigma_k>\sigma_{k+1}$, the optimal rank-$k$ approximation is unique.

\subsection{Low-Dimensional Residual Modulation}

We now apply Theorem~1 to the residual formulation used in our framework.
Let
\begin{equation*}
y=x+\delta,
\qquad
\delta=f(a;\,l),
\end{equation*}
where $x\in\mathbb{R}^{D}$ denotes the original RGB representation,
$a$ denotes the corresponding attribute representation,
$l$ denotes the language condition, and
$\delta\in\mathbb{R}^{D}$ denotes the resulting attribute-conditioned residual.

For spectral analysis, each $D$-dimensional residual token is treated as an individual sample, and residuals across observations and attribute branches are pooled together.
Denoting the resulting $n$ residual samples as
$\{\delta_i\}_{i=1}^{n}$, we first remove their common offset:
\begin{equation*}
\bar{\delta}=\frac{1}{n}\sum_{i=1}^{n}\delta_i,
\end{equation*}
and construct the centred residual matrix
\begin{equation*}
\widetilde{D}_{\delta}
=
\begin{bmatrix}
(\delta_1-\bar{\delta})^{\top}\\
\vdots\\
(\delta_n-\bar{\delta})^{\top}
\end{bmatrix}
\in\mathbb{R}^{n\times D}.
\end{equation*}
Let
\begin{equation*}
\widetilde{D}_{\delta}=U\Sigma V^{\top}
\end{equation*}
be its singular value decomposition, and let
$V_k=[v_1,\ldots,v_k]\in\mathbb{R}^{D\times k}$
contain the first $k$ right singular vectors.
We define the cumulative spectral energy retained by the top-$k$ components as
\begin{equation*}
\eta_k:=\frac{\sum_{j=1}^{k}\sigma_j^2}{\sum_{j=1}^{r}\sigma_j^2}.
\end{equation*}

According to Theorem 1, the optimal rank-$k$ approximation
$\widetilde{D}_{\delta}^{(k)}$ satisfies
\begin{equation*}
\frac{
\|\widetilde{D}_{\delta}
-\widetilde{D}_{\delta}^{(k)}\|_F^2
}{
\|\widetilde{D}_{\delta}\|_F^2
}
=
1-\eta_k.
\end{equation*}
Hence, if $\eta_k$ approaches $1$ for $k\ll D$, the centred residual variations can be accurately represented in the $k$-dimensional subspace spanned by $V_k$. Specifically, defining
\begin{equation*}
z_i=V_k^{\top}(\delta_i-\bar{\delta})
\in\mathbb{R}^{k},
\end{equation*}
the corresponding low-rank reconstruction is
\begin{equation*}
\delta_i
\approx
\bar{\delta}+V_k z_i.
\end{equation*}
Substituting this result into the original additive formulation gives
\begin{equation*}
y_i \approx x_i+\bar{\delta}+V_k z_i, \qquad k\ll D.
\end{equation*}

\noindent\textbf{Corollary 1.}
If the centred residual matrix satisfies $\eta_k\approx1$ for some $k\ll D$, then the residual variation admits an approximately
low-dimensional affine representation.
Accordingly, the additive formulation can be expressed as a shared residual offset $\bar{\delta}$ together with a low-dimensional sample-dependent component $V_kz_i$ added to the original RGB representation.

\subsection{Empirical Verification}

As shown in Table~\ref{tab:residual_geometry}, the residual representations broadly satisfy Corollary 1. Here, PCA is performed over the pooled token-level residual features
described above. For the RGB simulation tasks, the top 8 principal components retain \(95.7\%\) and \(94.0\%\) of the centred residual variation in Task VII and Task VII\(^{\mathrm{Table}}\), respectively, indicating that the unseen tabletop introduces little additional variation. The real-world thermal tasks are less concentrated, with the top 32 components retaining \(83.1\%\), \(84.4\%\), and \(89.6\%\) in Tasks II, IV, and VI. Given the token dimension \(D=2048\), most residual variation is still captured in a substantially lower-dimensional subspace. Equivalently, the empirical spectra support the approximation
\begin{equation*}
\delta_i
\approx
\bar{\delta}+V_k z_i,
\qquad
y_i
\approx
x_i+\bar{\delta}+V_k z_i,
\end{equation*}
with $k\ll D$.

We additionally measure the average pairwise cosine similarity between the residual branches. The similarity ranges from \(0.942\) to \(0.956\) for the thermal tasks and reaches \(0.999\) for the RGB tasks, showing strong directional alignment among attribute residuals. Together with the centred PCA results, these observations support interpreting the learned residuals as an approximately low-dimensional attribute-conditioned additive modulation of the original RGB representation.

\begin{table}[h]
\centering
\caption{
\textbf{Geometric analysis of attribute-conditioned residual representations.} $\eta_k$ denotes the cumulative centred PCA energy retained by the top-$k$ principal components.}
\label{tab:residual_geometry}
\begin{tabular}{lccccc}
\toprule
Task
& Avg. Cosine
& $\eta_4$
& $\eta_8$
& $\eta_{16}$
& $\eta_{32}$ \\
\midrule
Task II
& 0.942 & 54.2 & 66.1 & 74.5 & 83.1 \\
Task IV
& 0.947 & 58.3 & 69.3 & 76.5 & 84.4 \\
Task VI
& 0.956 & 69.6 & 79.6 & 84.5 & 89.6 \\
Task VII
& 0.999 & 88.1 & 95.7 & 98.4 & 99.5 \\
Task VII$^{\mathrm{Table}}$
& 0.999 & 84.8 & 94.0 & 97.6 & 99.1 \\
\bottomrule
\end{tabular}
\end{table}

\bibliographystyle{IEEEtran}
\bibliography{references}